\documentclass{article} % For LaTeX2e
\usepackage{iclr2027_conference,times}
\iclrfinaltrue
\usepackage{amsmath,amsfonts,bm}

\def\eqref#1{equation~\ref{#1}}
\def\1{\bm{1}}

\DeclareMathAlphabet{\mathsfit}{\encodingdefault}{\sfdefault}{m}{sl}
\SetMathAlphabet{\mathsfit}{bold}{\encodingdefault}{\sfdefault}{bx}{n}

\usepackage[most]{tcolorbox}
\usepackage{listings}
\usepackage{enumitem}

\usepackage{hyperref}
\usepackage{url}
\usepackage{caption} % 提供 \captionof
\usepackage{xcolor}

\usepackage{fontawesome5}
\definecolor{RefBlue}{HTML}{1F4E79}
\definecolor{CiteBlue}{HTML}{2A5DB0}
\definecolor{UrlCyan}{HTML}{007ACC}

\hypersetup{
    colorlinks=true,
    linkcolor=RefBlue,
    citecolor=CiteBlue,
    urlcolor=UrlCyan
}

\usepackage{setspace}
\usepackage{algorithm}
\usepackage{algpseudocode}

\usepackage{booktabs}
\usepackage{amsfonts}
\usepackage{multirow}
\usepackage{graphicx}
\usepackage[table]{xcolor}
\usepackage{makecell}

\definecolor{rawbg}{HTML}{FAFAFA}
\definecolor{rawborder}{HTML}{8A8F98}
\definecolor{rawtitle}{HTML}{ECEFF3}

\definecolor{enhbg}{HTML}{F7FBFF}
\definecolor{enhborder}{HTML}{4C78A8}
\definecolor{enhtitle}{HTML}{E4F0FA}

\definecolor{sysbg}{HTML}{FFFDF8}
\definecolor{sysborder}{HTML}{A77C2D}
\definecolor{systitle}{HTML}{FFF2D2}

\tcbset{
  promptstyle/.style={
    enhanced,
    coltitle=black,
    fonttitle=\bfseries,
    boxrule=1pt,
    arc=2.5mm,
    outer arc=2.5mm,
    left=3mm,
    right=3mm,
    top=1mm,
    bottom=1mm,
    toptitle=1.2mm,
    bottomtitle=1.2mm,
    before skip=0pt,
    after skip=3pt,
    segmentation hidden,
    lower separated=false,
    before upper={
      \setlength{\parindent}{0pt}
      \setlength{\parskip}{0pt}
    }
  }
}

\newtcolorbox{rawpromptbox}{
  promptstyle,
  colback=rawbg,
  colframe=rawborder,
  colbacktitle=rawtitle,
  title={Raw Prompt},
  after skip=2mm
}

\newtcolorbox{enhancedpromptbox}{
  promptstyle,
  colback=enhbg,
  colframe=enhborder,
  colbacktitle=enhtitle,
  title={Enhanced Prompt},
  after skip=2mm
}

\newtcblisting{systempromptbox}{
  enhanced,
  listing only,
  breakable,
  colback=sysbg,
  colframe=sysborder,
  coltitle=black,
  colbacktitle=systitle,
  title=\textbf{System Prompt},
  fonttitle=\bfseries,
  boxrule=1pt,
  arc=2.2mm,
  left=3mm,
  right=3mm,
  top=2mm,
  bottom=2mm,
  toptitle=1.2mm,
  bottomtitle=1.2mm,
  listing options={
    basicstyle=\ttfamily\small,
    columns=fullflexible,
    keepspaces=true,
    breaklines=true
  }
}

\definecolor{posgreen}{RGB}{0,150,0}
\definecolor{negred}{RGB}{180,0,80}

\usepackage[table]{xcolor}

\definecolor{lightgraycell}{RGB}{245,245,245}
\definecolor{lightbluecell}{RGB}{238,248,255}

\definecolor{gold}{RGB}{212,175,55}
\definecolor{silver}{RGB}{160,160,160}
\definecolor{bronze}{RGB}{205,127,50}

\definecolor{modelA}{HTML}{F6D3D0} % 淡珊瑚红
\definecolor{modelB}{HTML}{D9E2F7} % 淡蓝色
\definecolor{modelC}{HTML}{D7D6EF} % 淡靛蓝
\definecolor{ours}{HTML}{DCEEFF}
\usepackage{newfloat}
\usepackage{listings}

\usepackage{booktabs}
\usepackage{amsmath} 

\usepackage{amsthm}
\usepackage{mathtools}
\usepackage{amssymb}

\usepackage{makecell}
\title{PE-OPSD: Internalizing Prompt Enhancement into Flow-matching Models via On-Policy Self-Distillation}

\author{
Mingfeng Lin\textsuperscript{1}\thanks{Equal contribution.}
\quad
Chengfei Cai\textsuperscript{2}\footnotemark[1]
\quad
Lin Xu\textsuperscript{1}
\quad
Chengqian Ma\textsuperscript{3}
\quad
Yuxiang Wei\textsuperscript{4}
\quad
Liang Han\textsuperscript{1}\thanks{Corresponding author.}
\\
\textsuperscript{1}Harbin Institute of Technology (Shenzhen) \quad
\textsuperscript{2}Zhejiang University
\\
\textsuperscript{3}Peking University \quad
\textsuperscript{4}Harbin Institute of Technology
\\
\faGithub\ \textbf{Code: }
\href{https://github.com/sleepy1231/PE-OPSD}
{\texttt{https://github.com/sleepy1231/PE-OPSD}}
}

\begin{document}

\maketitle

\vspace{-10pt}
\begin{abstract}
Text-to-image users often provide concise and underspecified prompts, whereas generative models benefit from detailed textual conditions for reliable instruction following. Existing systems bridge this gap with Prompt Enhancers (PEs) that rewrite raw prompts at inference time, introducing additional latency and leaving prompt elaboration external to the generator. We instead view enhanced prompts as privileged training information and ask whether their benefits can be internalized. We propose Prompt-Enhanced On-Policy Self-Distillation (\textbf{PE-OPSD}) for text-to-image flow-matching models. During training, a raw-prompt student follows its own generation trajectory, while an enhanced-prompt teacher provides vector-field targets at the states visited by the student. This on-policy supervision distills the behavior induced by enhanced prompts into the raw-prompt student without requiring additional text--image pairs. At inference, both the PE and teacher are removed, and the student generates directly from raw prompts. Across multiple model families, PEs, and benchmarks, PE-OPSD achieves the strongest aggregate prompt fidelity among the evaluated baselines, yields positive aggregate visual appeal gains, and retains the base-model inference efficiency.
\end{abstract}

\section{Introduction}

User prompts are often concise descriptions rather than fully specified generation instructions~\citep{xie2023prompt,hahn2024proactive}. A prompt such as ``a girl reading under a tree'' specifies the main subject and scene while leaving pose, lighting, composition, and visual style open. Although these unspecified choices admit multiple valid realizations, explicitly instantiating them can help generative models produce visually coherent outputs and follow the stated constraints more reliably. Despite substantial progress in visual fidelity and sampling efficiency~\citep{zhang2023adding,yin2024dmd2,jiang2026dmdr}, reliable intent understanding remains difficult when prompts are short and underspecified~\citep{huang2026ape}. This creates a fundamental mismatch between concise user prompts and the detailed textual conditions preferred by generative models.

To bridge this gap, many industrial text-to-image systems employ a Prompt Enhancer (PE) before image generation~\citep{team2025hunyuanimage,zhao2026qwenimage}. Given a raw prompt, PE rewrites it into a longer description that makes implicit subject attributes, scene layouts and object relationships explicit. Prior works such as Promptist~\citep{promptist}, BeautifulPrompt~\citep{cao2023beautifulprompt}, and PromptEnhancer~\citep{wang2025promptenhancer} have shown that prompt rewriting can improve human preference, text--image relevance, attribute binding, and compositional relationships~\citep{ghosh2023geneval}. However, PE improves generation by modifying the input rather than improving the generator itself: prompt enhancement remains delegated to an additional inference-time module.

This deployment paradigm has practical and conceptual limitations. Practically, prompt rewriting introduces extra latency and computation, while longer prompts increase text-processing and conditioning overhead~\citep{wang2023reprompt}. Conceptually, an ideal generator should not require an external rewriter for every generation. It is therefore desirable for a generator to internalize the generation behavior induced by PEs and generate high-quality images directly from concise prompts.

This raises a natural question: \textbf{\emph{Can a generator capture the generation benefits elicited by prompt enhancement while receiving only the raw prompt at inference time?}} Existing approaches exploit richer textual conditions mainly in two ways. Some replace original text conditions with expanded captions during training~\citep{betker2023dalle}, which improves training data quality but does not directly teach the model to recover enhanced-prompt behavior from raw prompts. Others keep the generator fixed and continuously invoke PE at inference time~\citep{cao2023beautifulprompt,manas2024improving,wang2025promptenhancer}, which is effective but retains the deployment cost. Thus, less attention has been paid to internalizing the benefit of prompt enhancement into the generator itself.

\begin{figure}[t]
    \centering
    \includegraphics[width=1.0\linewidth]{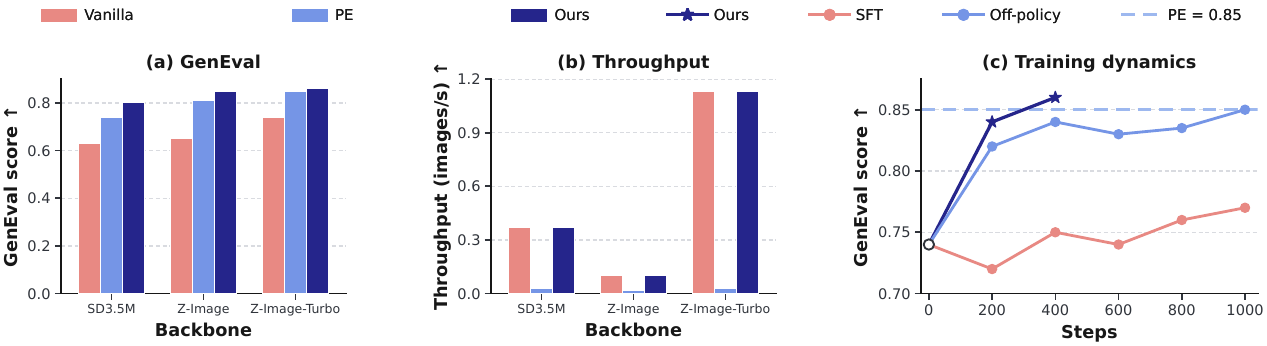}
    \caption{Comparison of generation quality, throughput, and training dynamics. PE-OPSD achieves higher GenEval scores than both Vanilla and PE while preserving the inference efficiency. Moreover, PE-OPSD achieves faster convergence and greater performance gains than SFT and off-policy distillation. Training dynamics is reported in \texttt{Z-Image-Turbo} training.}
    \label{fig:dynamics}
    \vspace{-10pt}
\end{figure}

In this work, we reinterpret prompt enhancement as a form of \emph{privileged information}. Unlike conventional privileged information from additional modalities, an enhanced prompt elaborates the raw prompt while preserving its explicit semantic constraints. Crucially, we do not feed this privileged text to the student. Instead, it is used only to define a better-informed teacher condition, while the student must learn to reproduce the corresponding generation behavior from the raw prompt alone.

Building on this perspective, we propose Prompt-Enhanced On-Policy Self-Distillation (PE-OPSD) for text-to-image flow-matching models. Given a raw prompt $p$, the PE produces an enhanced prompt $p^+$. The student is conditioned only on $p$, while the teacher receives $p^+$ as privileged information. During training, the student follows its own vector field to generate on-policy flow trajectories. At the states visited by the student, the teacher provides target vector fields conditioned on $p^+$. By matching these targets under the raw-prompt condition, the student learns to approximate the generation behavior induced by enhanced prompts. Unlike supervised finetuning (SFT), PE-OPSD does not require constructing additional text-image pairs. Unlike off-policy distillation based on teacher trajectories, it supervises the student on its own raw-prompt trajectories, better aligning the training signal with the states encountered during raw-prompt generation. In this way, PE-OPSD converts PE from an inference-time module into a training-time supervision signal on flow dynamics. After training, both PE and teacher are removed, enabling direct generation from raw prompts without inference-time prompt rewriting.

Our contributions are summarized threefold:

\begin{itemize}
    \item We reinterpret prompt enhancement as privileged information for text-to-image generation, studying how the benefits of enhanced prompts can be transferred to a model that observes only raw prompts at inference time.
    \item We propose Prompt-Enhanced On-Policy Self-Distillation (PE-OPSD) for text-to-image flow-matching models, where an enhanced-prompt teacher provides dense vector-field supervision on states visited by a raw-prompt student, thereby distilling enhanced-prompt generation behavior into a model conditioned only on raw prompts.
    \item We show that PE-OPSD outperforms inference-time PE, SFT, and off-policy distillation in aggregate prompt fidelity on all three main backbones, while retaining positive aggregate visual-appeal changes and base-model inference latency.
\end{itemize}

\section{Related Work}

\paragraph{Prompt enhancement for text-to-image generation.}
Prompt enhancement aims to bridge the gap between underspecified user prompts and the detailed textual conditions under which text-to-image models more reliably satisfy explicit prompt constraints and produce visually coherent outputs. Promptist~\citep{promptist} learns model-preferred prompts with reinforcement learning, BeautifulPrompt~\citep{cao2023beautifulprompt} trains a PE from low- and high-quality prompt pairs with visual feedback, and PromptEnhancer~\citep{wang2025promptenhancer} further improves PE through chain-of-thought reasoning and fine-grained reward signals. However, these improvements are largely achieved outside the generator by training an external PE. In contrast, our work uses enhanced prompts only as training-time privileged information and distills their effect into the generator, enabling inference directly from raw prompts.
% Other prompt-side approaches refine textual conditions through automatic prompt editing or dynamic token-level control, while recaptioning improves generation by replacing underspecified captions with more descriptive textual supervision.

\paragraph{On-policy distillation.}
On-policy distillation (OPD)~\citep{agarwal2024gkd,gu2024minillm} mitigates train--inference mismatch by supervising the student on samples generated from its own current policy rather than from a fixed offline distribution. On-policy self-distillation (OPSD)~\citep{zhao2026opsd} extends this idea by constructing asymmetric teacher and student views from the same base model, reducing the need for a separate stronger teacher. In recent visual and multimodal methods~\citep{bi2026opdv,liu2026opsdv,yuan2026visionopd}, this asymmetry is often induced by privileged information available only to the teacher during training, such as cropped regions, higher-resolution inputs, or visual reasoning traces. D-OPSD~\citep{jiang2026dopsd} further introduces this paradigm to step-distilled diffusion models, but relies on paired image--text data and a generator capable of accepting image-conditioned inputs. PE-OPSD differs from these methods by using enhanced prompts only as training-time privileged information, thereby transferring their effect to a raw-prompt flow-matching model without requiring external prompt enhancement at inference.

\section{Methodology}

\begin{figure}[t]
    \centering
    \begin{minipage}[c]{0.63\linewidth}
        \vspace{0pt}
        \centering
        \includegraphics[width=0.98\linewidth]{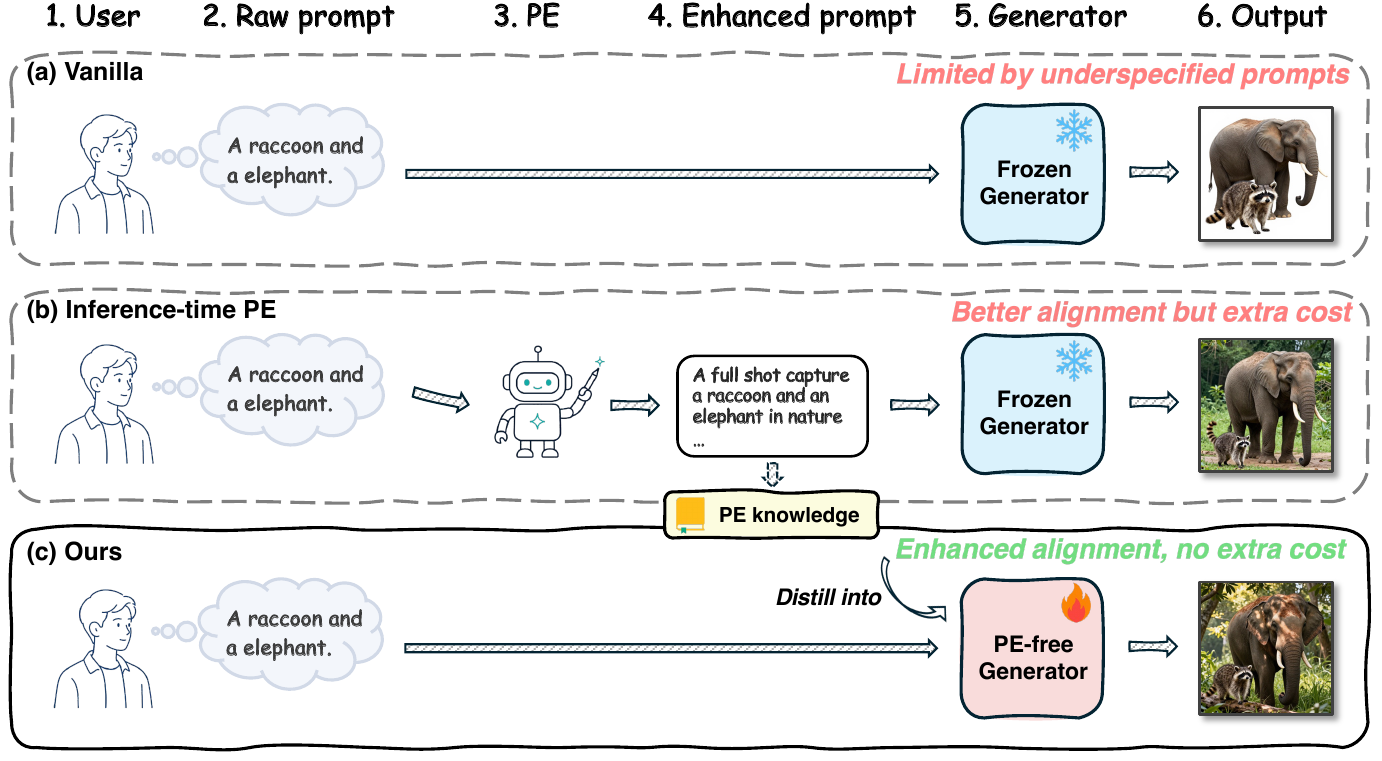}
    \end{minipage}\hfill
    \begin{minipage}[c]{0.35\linewidth}
        \vspace{10pt}
        \captionof{figure}{\textbf{Three different uses of prompt conditioning.} (a)\;Direct generation from raw prompts results in limited prompt alignment; (b)\;Inference-time PE improves prompt alignment by rewriting raw prompts, but introduces additional computational overhead; (c)\;Our PE-OPSD internalizes the PE knowledge into the generator, improving alignment without additional inference cost.}
        \label{fig:comp}
    \end{minipage}
    \vspace{-10pt}
\end{figure}

\subsection{Preliminaries}
\paragraph{On-policy distillation.}
Let $f_\theta$ denote a student model and $f_{\phi}$ denote a teacher model. Conventional knowledge distillation~\citep{hinton2015distilling,beyer2022knowledge} supervises the student on examples drawn from a fixed data distribution, which may differ from the states encountered by the student at inference. On-policy distillation instead evaluates the teacher on samples generated by the current student distribution. In a generic form, the objective can be written as
\[
\mathcal{L}_{\mathrm{OPD}}(\theta)
=
\mathbb{E}_{z \sim q_\theta(\cdot \mid c)}
\left[
D_{\mathrm{KL}}\big(
f_\theta(z,c),
f_{\phi}(z,c)
\big)
\right],
\]
where $c$ denotes the conditioning input, $q_\theta(\cdot \mid c)$ is the distribution induced by the current student.

\paragraph{On-policy self-distillation.}
On-policy self-distillation further removes the need for a separate teacher model by deriving teacher and student signals from asymmetric views of the same base model. Let $c_s$ denote the student condition and $c_t$ denote a stronger teacher condition, where $c_t$ is the privileged information available only during training. The corresponding objective is
\[
\mathcal{L}_{\mathrm{OPSD}}(\theta)
=
\mathbb{E}_{z \sim q_\theta(\cdot \mid c_s)}
\left[
D_{\mathrm{KL}}\big(
f_\theta(z,c_s),
f_{\bar{\theta}}(z,c_t)
\big)
\right],
\]
where $f_{\bar{\theta}}$ denotes the teacher model, which may be a stop-gradient version of the base model or an exponential-moving-average (EMA) \citep{morales2024exponential} copy of the student model.

\subsection{From Prompt Enhancement to Privileged Information}

\begin{figure*}[h]
    \centering
    \begin{minipage}[t]{0.46\textwidth}
        \vspace{0pt}
        \centering
        \includegraphics[width=\linewidth]{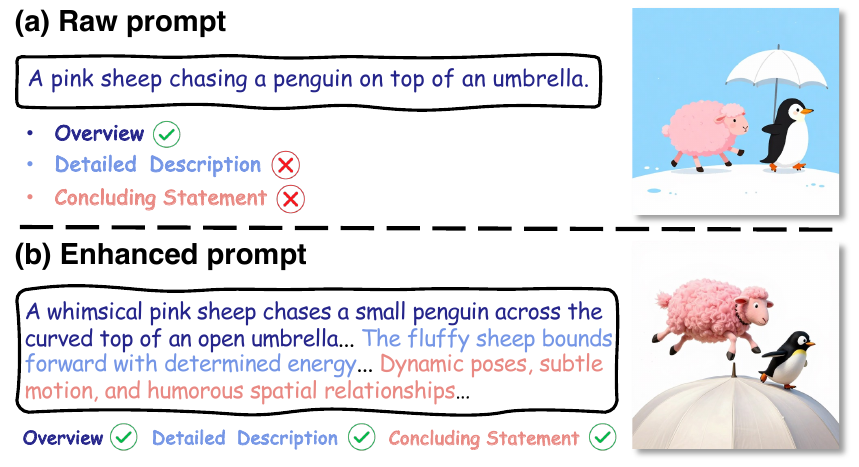}
        \vspace{-15pt}
        \captionof{figure}{Comparison of raw and enhanced prompt, with the enhanced prompt providing denser and more detailed textual information.}
        \label{fig:insight}
    \end{minipage}\hfill
    \begin{minipage}[t]{0.52\textwidth}
        \vspace{0pt}
        \centering
        \captionof{table}{Comparison of the Base and inference-time PE on GenEval and GenEval2. $^\dagger$denotes \texttt{GPT-5.6 Sol} as PE. $^\ddagger$denotes \texttt{PromptEnhancer} as PE.}
        \vspace{-5pt}
        \setlength{\tabcolsep}{1mm}
        {
        \scriptsize
        \renewcommand{\arraystretch}{1}
        \begin{tabular}{lcc|ccc}
            \toprule
            \multirow{2}{*}{\textbf{Method}}
            & \multicolumn{2}{c|}{\textbf{GenEval Task}}
            & \multicolumn{3}{c}{\textbf{GenEval2 Task}} \\
            \cmidrule(lr){2-3}
            \cmidrule(lr){4-6}
            & \textbf{Token Len.} & \textbf{GE}
            & \textbf{Token Len.} & \textbf{GE2$_{\texttt{AM}}$} & \textbf{GE2$_{\texttt{GM}}$} \\
            \midrule
            Base & 11.02 & 0.737 & 7.92 & 0.783 & 0.341 \\
            \midrule
            PE-GPT$^\dagger$ & 152.61 & 0.850 & 135.11 & 0.843 & 0.479 \\
            \rowcolor{lightbluecell}
            \quad $\Delta$ (vs Base)  & \color{posgreen}{13.9$\times$} & \color{posgreen}{+0.113} & \color{posgreen}{17.1$\times$} & \color{posgreen}{+0.060} & \color{posgreen}{+0.138} \\
            PE-7B$^\ddagger$ & 198.29 & 0.746 & 170.66 & 0.802 & 0.383 \\
            \rowcolor{lightbluecell}
            \quad $\Delta$ (vs Base)  & \color{posgreen}{18.0$\times$} & \color{posgreen}{+0.009} & \color{posgreen}{21.5$\times$} & \color{posgreen}{+0.019} & \color{posgreen}{+0.042} \\
            PE-32B$^\ddagger$ & 155.82 & 0.859 & 148.66 & 0.849 & 0.492 \\
            \rowcolor{lightbluecell}
            \quad $\Delta$ (vs Base)  & \color{posgreen}{14.1$\times$} & \color{posgreen}{+0.122} & \color{posgreen}{18.8$\times$} & \color{posgreen}{+0.066} & \color{posgreen}{+0.151} \\
            \bottomrule
        \end{tabular}}
        \label{tab:pe_result}
    \end{minipage}
    \vspace{-5pt}
\end{figure*}

\paragraph{Prompt enhancement elicits conditional capability.}
Let $p$ denote a raw user prompt and let $E$ be a prompt enhancer that produces an enhanced prompt $p^{+}=E(p)$. Figure~\ref{fig:insight} and Table~\ref{tab:pe_result} compare these conditions using the same frozen Z-Image-Turbo generator~\citep{cai2025zimage}. Across the tested PEs, enhanced prompts improve the reported compositional scores while substantially increasing prompt length. By adding descriptive details to \(p\) while preserving its explicit constraints, PE provides a richer condition for generation.

We view \(p^{+}\) as a PE-selected elaboration of \(p\), intended to preserve the explicit semantics of \(p\) while adding one plausible realization of otherwise unspecified attributes, composition, and scene details. Because prompt enhancement changes only the conditioning input while keeping the generator fixed~\citep{promptist,wang2025promptenhancer}, the performance gap between \(p\) and \(p^{+}\) shows that the pretrained generator can better realize the requested content when conditioned on a more detailed description. This does not imply that the generator can recover the missing details from \(p\) alone. Instead, it shows that the improved generation behavior is attainable under enhanced prompts. We therefore seek to internalize this enhanced-prompt behavior into a model conditioned only on $p$.

\paragraph{Enhanced prompts as native textual privileged information.}
We cast this objective through the lens of learning with privileged information. During training, $p^{+}$ provides the teacher with a richer PE-generated condition associated with the same raw prompt, while the student remains conditioned only on $p$; at deployment, $p^{+}$ is unavailable. Importantly, the teacher's advantage arises from asymmetric conditioning rather than greater model capacity. Prompt enhancement therefore constitutes a native form of textual privileged information: both $p$ and $p^{+}$ are processed by the same architecture through its existing text-conditioning pathway, without introducing an auxiliary modality.

This differs from image-privileged formulations such as D-OPSD~\citep{jiang2026dopsd}, which construct the teacher condition by jointly encoding the prompt and a paired target image with a multimodal encoder. PE-OPSD instead derives the privileged condition from prompt enhancement alone, requiring neither paired target images nor modifications to the model's conditioning interface. Crucially, our goal is not to reconstruct $p^{+}$ or imitate the PE itself, but to distill the generation dynamics elicited by $p^{+}$ into a student that observes only $p$. We next formalize this principle for flow-matching generators.

\subsection{PE-OPSD: Prompt-Enhanced On-Policy Self-Distillation}

\begin{figure}[t]
    \centering
    \includegraphics[width=0.95\linewidth]{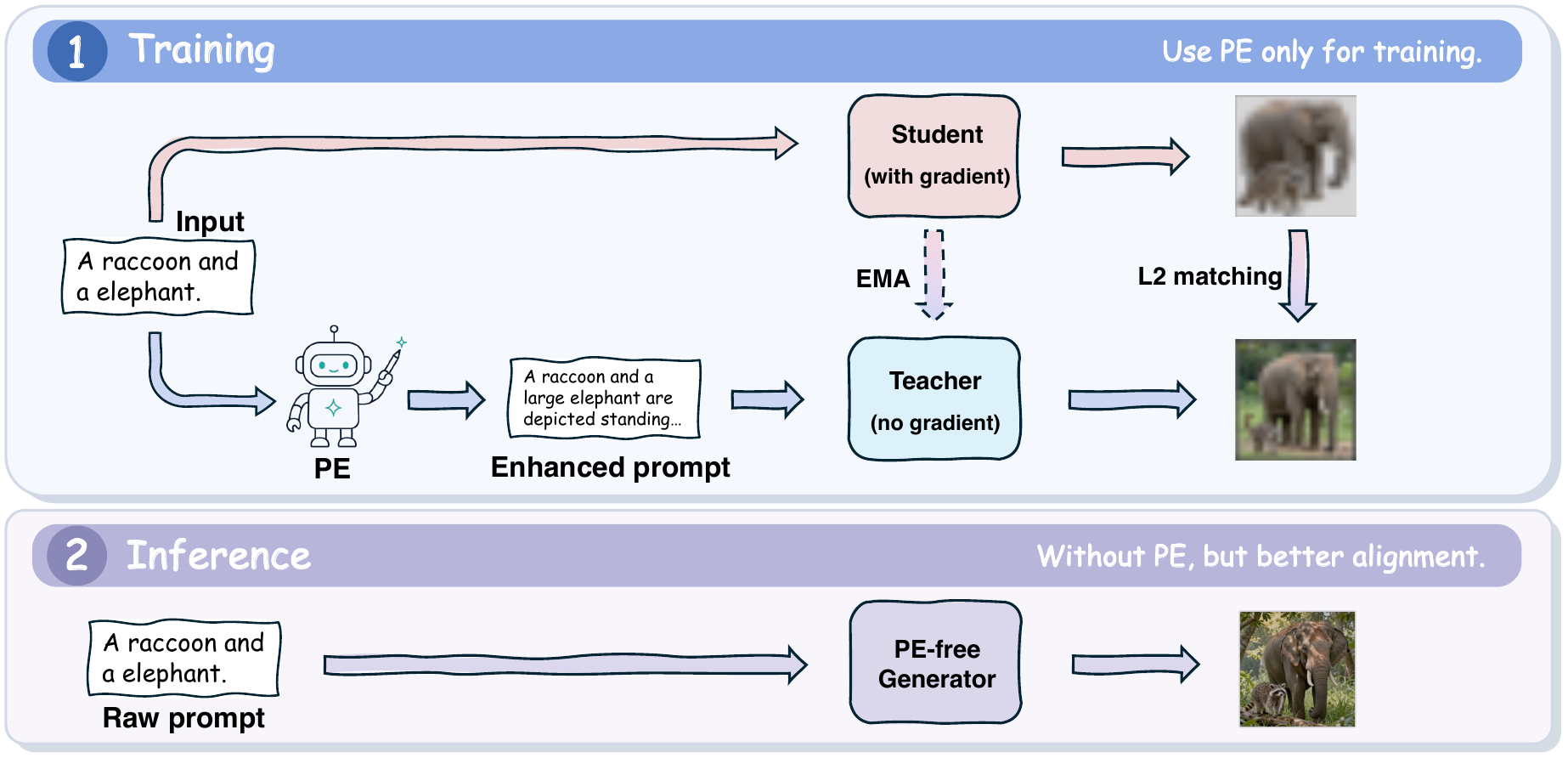}
    \caption{\textbf{Overview of PE-OPSD.} At training, the student generates rollouts from raw prompts, while an EMA teacher provides enhanced-prompt supervision on the same rollout states. At inference, the trained student generates directly from raw prompts without inference-time PE.}
    \label{fig:summary}
    \vspace{-10pt}
\end{figure}

\paragraph{Overview.}
Figure~\ref{fig:summary} illustrates the training and inference pipelines of PE-OPSD. Let $\mathcal{D}=\{(p,p^{+})\}$ denote the training prompt pairs set, where $p^{+}=E(p)$ is obtained through PE. The student and teacher share the same architecture and are initialized from the same pretrained parameters, $\theta=\bar{\theta}=\theta_{\mathrm{pre}}$. During training, the student generates trajectories conditioned only on $p$, while the teacher provides enhanced-prompt supervision on the states visited by the student. After training, the behavior induced by enhanced prompts is internalized into the generator, enabling direct raw-prompt generation without invoking PE at inference time.

\paragraph{On-policy sampling under raw prompts.}
We collect training states from the current student's own generation process. Let $1=t_K>t_{K-1}>\cdots>t_0=0$ denote the discrete denoising schedule, and let $\Delta t_k=t_k-t_{k-1}>0$. Starting from $x_{t_K}\sim\mathcal{N}(0,I)$, the student follows the Euler updates
\begin{equation}
    x_{t_{k-1}}
    =
    x_{t_k}-\Delta t_k v_{\theta}(x_{t_k},t_k,p),
    \qquad k=K,\ldots,1.
    \label{eq:pe_opsd_rollout}
\end{equation}

We supervise the student at the visited states $\tau=\{x_{t_k}\}_{k=1}^{K}$. Because these states are generated by the current student under the raw condition $p$, $\tau$ follows the state induced by the current student under the training sampler. In contrast, a trajectory generated by the teacher under $p^{+}$ would follow a different state distribution and thus provide off-policy supervision for the raw-prompt student.

\paragraph{Distillation from enhanced-prompt supervision.}
At each student-visited state $x_{t_k}$, the student and teacher are evaluated at the same state and time but under different prompts:
\begin{equation}
    v_k^{S}=v_{\theta}(x_{t_k},t_k,p), \qquad
    v_k^{T}=v_{\bar{\theta}}(x_{t_k},t_k,p^{+}).
    \label{eq:pe_opsd_predictions}
\end{equation}

% Both predictions use the same latent state and timestep. The enhanced prompt therefore changes the teacher's supervision without changing the states on which the student is trained.

% For Gaussian transitions with shared covariance $\sigma_k^2 I$, transition KL reduces to a scaled squared distance between their means~\citep{li2026diffusionopd}. We use this connection to motivate direct squared-error supervision for deterministic flow sampling.
% \begin{equation}
%     D_{\mathrm{KL}}\!\left(
%         \mathcal{N}(\mu_k^{S},\sigma_k^2 I)
%         \,\middle\|\,
%         \mathcal{N}(\mu_k^{T},\sigma_k^2 I)
%     \right)
%     =
%     \frac{
%         \|\mu_k^{S}-\mu_k^{T}\|_2^2
%     }{2\sigma_k^2},
%     \qquad \sigma_k^2>0.
%     \label{eq:pe_opsd_transition_kl}
% \end{equation}
Because PE-OPSD operates with deterministic flow trajectories, we directly regress the teacher's vector field rather than introducing a stochastic transition kernel. We consider matching velocities $v$, one-step transitions $\mu$, or predicted clean latents $\hat{x}_0$. Under the linear flow interpolation $x_t=(1-t)x_0+t\epsilon$~\citep{lipman2022flow}, with $\epsilon\sim\mathcal{N}(0,I)$, the latter two targets are
\begin{equation}
    \mu_k^{b}=x_{t_k}-\Delta t_k\,v_k^{b},
    \qquad
    \hat{x}_{0,k}^{b}=x_{t_k}-t_k\,v_k^{b},
    \qquad b\in\{S,T\}.
    \label{eq:pe_opsd_parameterizations}
\end{equation}

Since both $\mu_k$ and $\hat{x}_{0,k}$ share the same state $x_{t_k}$, their squared-error objectives reduce to a weighted velocity mismatch. We therefore express all three variants using a unified objective:
\begin{equation}
    \mathcal{L}_{\mathrm{PE\text{-}OPSD}}(\theta;\bar{\theta})
    =
    \mathbb{E}_{\substack{
        (p,p^{+})\sim\mathcal{D}\\
        z\sim\mathcal{N}(0,I)
    }}
    \left[
        \frac{1}{K}
        \sum_{k=1}^{K}
        \omega(t_k)
        \left\|v_k^{S}-\operatorname{sg}\!\left[v_k^{T}\right]\right\|_2^2
    \right],
    \label{eq:pe_opsd_loss}
\end{equation}
where $\omega_v(t_k)=1$, $\omega_{x_0}(t_k)=t_k^2$, and $\omega_{\mu}(t_k)=(t_{k-1}-t_k)^2=(\Delta t_k)^2$, respectively. $\operatorname{sg}\!\left[\cdot \right]$ denotes stop-gradient. We refer to these variants as $v{\mathrm{-loss}}$, $x_0\mathrm{-loss}$, and $\mu\mathrm{-loss}$. The three variants share the same pointwise optimum in velocity space but differ in timestep weighting. Appendix~\ref{app:kl_to_ode} provides a trajectory-level KL interpretation and its connection to our deterministic matching objectives.

During optimization, the teacher predictions are detached, and gradients pass only through the student predictions in Equation~\ref{eq:pe_opsd_predictions}. After updating the student from $\theta_n$ to $\theta_{n+1}$, we update the teacher through EMA as
\begin{equation}
    \bar{\theta}_{n+1}\leftarrow
    \gamma\,\bar{\theta}_{n}+(1-\gamma)\,\theta_{n+1},
    \qquad 0\leq\gamma<1,
    \label{eq:pe_opsd_ema}
\end{equation}
where $\gamma$ is the EMA decay rate. This provides a temporally smoothed teacher that evolves with the student while retaining enhanced-prompt conditioning.

\paragraph{Training recipe.}
Algorithm~\ref{alg:pe_opsd} summarizes the training procedure. We first construct prompt pairs set $\mathcal{D}=\{(p,p^{+})\}$ by applying the off-the-shelf PE offline, avoiding PE calls during training. Each iteration rollouts the student using the raw prompt $p$, and evaluates the enhanced-prompt teacher at the visited states. The student is optimized to match the stop-gradient teacher velocities, after which the teacher is updated by EMA. After training, only the student is retained and generation proceeds directly from raw prompts without invoking the PE or teacher.

\vspace{-5pt}
\begin{algorithm}[h]
    \caption{PE-OPSD Training}
    \label{alg:pe_opsd}
    \begingroup
    % \small
    \setstretch{0.85}
    \begin{algorithmic}[1]
        \Require Prompt pairs set $\mathcal{D}=\{(p,p^{+})\}$; pretrained model $\theta_{\mathrm{pre}}$; schedule $\{t_0,\cdots,t_K\}$; EMA decay rate $\gamma$
        \State $\theta\gets\theta_{\mathrm{pre}}$,
               $\bar{\theta}\gets\theta_{\mathrm{pre}}$
        \For{each training iteration}
            \State Sample a mini-batch $(p,p^{+})\sim\mathcal{D}$ and $x_{t_K}\sim\mathcal{N}(0,I)$
            \State Initialize $\mathcal{L}\gets 0$
            \For{each $k=K,\ldots,1$}
                % \State $v^{S}\gets v_{\theta}(x_{t_k},t_k,p)$
                % \State $v^{T}\gets v_{\bar{\theta}}(x_{t_k},t_k,p^{+})$
                \State $\mathcal{L} \gets \mathcal{L}+\frac{\omega(t_k)}{K}
                       \|v_{\theta}(x_{t_k},t_k,p)-\operatorname{sg}[v_{\bar{\theta}}(x_{t_k},t_k,p^{+})]\|_2^2$ \Comment{$\mu\mathrm{-loss}$ by default}
                \If{$k>1$}
                    \State $x_{t_{k-1}}\leftarrow \operatorname{sg}[x_{t_k}-\Delta t_k v_{\theta}(x_{t_k},t_k,p)]$ \Comment{Rollout for one step}
                \EndIf
            \EndFor
            \State $\theta\gets
                   \operatorname{OptimizerStep}
                   (\theta,\nabla_{\theta}\mathcal{L})$
            \State $\bar{\theta}\gets
                   \gamma\bar{\theta}+(1-\gamma)\theta$ \Comment{EMA update}
        \EndFor
        \State \Return $\theta$
    \end{algorithmic}
    \endgroup
\end{algorithm}
\vspace{-10pt}

\section{Experiments}
\subsection{Experimental Setups}
\paragraph{Models and baselines.} Our main experiments use SD3.5-M~\citep{esser2024sd}, Z-Image, and Z-Image-Turbo, comparing PE-OPSD against the unmodified model (Base), inference-time prompt enhancement (Base+PE), supervised fine-tuning (SFT), and off-policy distillation. For SFT, we first generate teacher images conditioned on enhanced prompts offline and then fine-tune the student on the images. Off-policy distillation uses the same teacher and distillation objective as PE-OPSD but collects training states from enhanced-prompt teacher trajectories rather than raw-prompt student trajectories. For fairness, all trainable methods use the same initialization, training data, optimization configuration, and number of training steps. We further evaluate FLUX.2-klein-base, FLUX.2-klein~\citep{blackforestlabs2026flux2klein}, and QwenImage-2512~\citep{zhao2026qwenimage} to assess scalability. We use \texttt{GPT-5.6 Sol} as the default PE and evaluate PE generalization on Z-Image with \texttt{PromptEnhancer-7B/32B}~\citep{wang2025promptenhancer}.

\paragraph{Benchmarks and metrics.} We evaluate compositional generation on GenEval (GE,~\citet{ghosh2023geneval}) and GenEval2 (GE2,~\citet{kamath2025geneval2}), reporting GE and GE2$_{\texttt{GM/AM}}$, alongside CLIP score~\citep{hessel2021clipscore}, PickScore~\citep{kirstain2023pick}, and aesthetics~\citep{schuhmann2022aesthetics} on both benchmarks. We summarize improvement over Base as
\begin{equation}
\mathbb{I}_{c}=\frac{100\%}{|\mathcal{M}_{c}|}\sum_{m\in\mathcal{M}_{c}}\left(\frac{\mathrm{Score}_{\mathrm{method}}-\mathrm{Score}_{\mathrm{Base}}}{\mathrm{Score}_{\mathrm{Base}}}\right), \quad c=\{\mathrm{PF},\;\mathrm{VA}\}
\end{equation}
where $\mathcal{M}_{\mathrm{PF}}$ contains GE, GE2$_{\texttt{GM}}$, GE2$_{\texttt{AM}}$, and CLIP scores on both benchmarks, while $\mathcal{M}_{\mathrm{VA}}$ contains PickScore and aesthetics on both benchmarks. These indices measure \textbf{P}rompt \textbf{F}idelity and \textbf{V}isual \textbf{A}ppeal improvements, respectively. Out-of-domain evaluation uses DPG-bench~\citep{hu2024dpg}, T2I-CompBench++~\citep{huang2025t2i}, and EvalMuse~\citep{han2026evalmuse}.

\paragraph{Configuration.}
For GenEval, we adopt the training and evaluation splits released by Flow-GRPO~\citep{liu2026flow}. For GenEval2, we train on the official 20K synthetic prompts and evaluate on the 800 officially released prompts. All models are trained on the same mixture of the GenEval and GenEval2 training sets. To ensure a fair comparison, we use a unified training recipe across all models and methods. Full configuration details and experimental setups are provided in Appendix~\ref{app:exp}.

\subsection{Main Experiments}
\paragraph{Main results.} Table~\ref{tab:main_results} shows that PE-OPSD consistently achieves the best prompt fidelity across all three backbones. It obtains the highest GE, GE2$_{\texttt{GM}}$, and $\mathbb{I}_{\mathrm{PF}}$, outperforming SFT, off-policy distillation, and even Base+PE, which retains the PE at inference. Relative to Base, PE-OPSD improves aggregate prompt fidelity by \(13.35\%\), \(16.07\%\), and \(15.22\%\) on SD3.5-M, Z-Image, and Z-Image-Turbo, respectively. Meanwhile, $\mathbb{I}_{\mathrm{VA}}$ remains positive on every backbone, indicating that PE-OPSD improves prompt fidelity without degrading aggregate visual appeal.

\begin{table*}[t]
\centering
\setlength{\tabcolsep}{1.25mm}
\caption{\textbf{Main results across different models.} We use \texttt{GPT-5.6 Sol} as PE. PickScore is normalized by 26; Aes. denotes aesthetics; \textbf{Bold:} best; \underline{Underlined:} second-best.}
\vspace{-8pt}
{
\scriptsize
\renewcommand{\arraystretch}{0.95}
\begin{tabular}{lccccccccccc}
\toprule
\multirow{2}{*}[-0.8ex]{\textbf{Method}}
& \multicolumn{4}{c}{\textbf{GenEval (GE) Task}}
& \multicolumn{5}{c}{\textbf{GenEval2 (GE2) Task}} \\
\cmidrule(lr){2-5}
\cmidrule(lr){6-10}
& \textbf{GE}
& \textbf{PickScore}
& \textbf{CLIP}
& \textbf{Aes.}
& \textbf{GE2$_{\texttt{GM}}$}
& \textbf{GE2$_{\texttt{AM}}$}
& \textbf{PickScore}
& \textbf{CLIP}
& \textbf{Aes.} 
& \multirow{2}{*}[+3.0ex]{\textbf{$\mathbb{I}_{\mathrm{PF}}$}}
& \multirow{2}{*}[+3.0ex]{\textbf{$\mathbb{I}_{\mathrm{VA}}$}}
\\
\midrule
SD3.5-L
& 0.604 & 0.862 & 0.288 & 5.286 & 0.213 & 0.660 & 0.869 & 0.315 & 5.521 & - & - \\
FLUX.1-dev
& 0.618 & 0.899 & 0.280 & 5.624 & 0.179 & 0.643 & 0.895 & 0.298 & 5.909 & - & - \\
\midrule
\rowcolor{lightgraycell}
\multicolumn{12}{c}{\textbf{\texttt{SD3.5-M (2.5B)}}} \\
% \midrule
Base
& 0.628 & 0.878 & 0.289 & 5.319 & 0.176 & 0.633 & 0.882 & 0.313 & 5.579 & 0.00\% & 0.00\%\\
Base+PE
& 0.743 & \underline{0.892} & 0.291 & \underline{5.439} & 0.220 & \underline{0.680} & 0.887 & 0.312 & 5.678 & 10.22\% & 1.55\%\\
SFT
& 0.718 & 0.889 & \underline{0.295} & \textbf{5.462} & 0.207 & 0.656 & 0.886 & 0.310 & \underline{5.702} & 7.34\% & \underline{1.65\%}\\
Off-Policy Distillation
& \underline{0.770} & \underline{0.892} & \textbf{0.296} & 5.407 & \underline{0.223} & 0.671 & \underline{0.890} & \underline{0.316} & 5.567 & \underline{11.74\%} & 0.99\% \\
\rowcolor{lightbluecell}
PE-OPSD (Ours)
& \textbf{0.797} & \textbf{0.896} & \underline{0.295} & 5.401 & \textbf{0.226} & \textbf{0.682} & \textbf{0.893} & \textbf{0.318} & \textbf{5.708} & \textbf{13.35\%} & \textbf{1.79\%} \\
\rowcolor{lightbluecell}
\qquad $\Delta$ (vs Base) 
& \color{posgreen}{+0.169} & \color{posgreen}{+0.018} & \color{posgreen}{+0.006} & \color{posgreen}{+0.082} & \color{posgreen}{+0.050} & \color{posgreen}{+0.049} & \color{posgreen}{+0.011} & \color{posgreen}{+0.005} & \color{posgreen}{+0.129} & \color{posgreen}{+13.35\%} & \color{posgreen}{+1.79\%}\\
\midrule
\rowcolor{lightgraycell}
\multicolumn{12}{c}{\textbf{\texttt{Z-Image (6B)}}} \\
% \midrule
Base
& 0.650 & 0.875 & 0.288 & 5.288 & 0.306 & 0.761 & 0.871 & 0.319 & 5.421 & 0.00\% & 0.00\%  \\
Base+PE
& 0.810 & 0.899 & 0.299 & \textbf{5.433} & \underline{0.404} & 0.827 & 0.893 & 0.326 & \underline{5.638} & 14.27\% & 3.00\%\\
SFT
& 0.821 & 0.893 & \textbf{0.302} & 5.426 & 0.403 & \textbf{0.834} & 0.887 & 0.326 & 5.466 & \underline{14.93\%} & 1.83\% \\
Off-Policy Distillation
& \underline{0.824} & \underline{0.901} & \underline{0.300} & \underline{5.432} & 0.395 & 0.828 & \underline{0.897} & \underline{0.327} & 5.629 & 14.27\% & \underline{3.12\%} \\
\rowcolor{lightbluecell}
PE-OPSD (Ours)
& \textbf{0.846} & \textbf{0.902} & \textbf{0.302} & 5.423 & \textbf{0.406} & \underline{0.831} & \textbf{0.900} & \textbf{0.330} & \textbf{5.650} & \textbf{16.07\%} & \textbf{3.30\%} \\
\rowcolor{lightbluecell}
\qquad $\Delta$ (vs Base) 
& \color{posgreen}{+0.196} & \color{posgreen}{+0.027} & \color{posgreen}{+0.014} & \color{posgreen}{+0.135} & \color{posgreen}{+0.100} & \color{posgreen}{+0.070} & \color{posgreen}{+0.029} & \color{posgreen}{+0.011} & \color{posgreen}{+0.229} & \color{posgreen}{+16.07\%} & \color{posgreen}{+3.30\%}\\
\midrule
\rowcolor{lightgraycell}
\multicolumn{12}{c}{\textbf{\texttt{Z-Image-Turbo (6B)}}} \\
% \midrule
Base
& 0.737 & 0.908 & 0.291 & \underline{5.289} & 0.341 & 0.783 & 0.901 & 0.320 & 5.436 & 0.00\% & 0.00\%\\
Base+PE
& 0.850 & \underline{0.918} & \textbf{0.298} & \textbf{5.306} & \underline{0.479} & \underline{0.843} & \underline{0.911} & \underline{0.325} & \textbf{5.561} & \underline{13.49\%} & \textbf{1.21\%}\\
SFT
& 0.766 & 0.893 & 0.295 & 5.218 & 0.413 & 0.788 & 0.884 & 0.317 & 5.398 & 5.22\% & -1.40\%\\
Off-Policy Distillation
& \underline{0.851} & 0.916 & \underline{0.296} & 5.251 & 0.437 & 0.821 & 0.910 & \underline{0.325} & \textbf{5.561} & 10.35\% & 0.87\% \\
\rowcolor{lightbluecell}
PE-OPSD (Ours)
& \textbf{0.863} & \textbf{0.919} & \textbf{0.298} & 5.274 & \textbf{0.501} & \textbf{0.844} & \textbf{0.913} & \textbf{0.326} & \underline{5.547} & \textbf{15.22\%} & \underline{1.08\%}\\
\rowcolor{lightbluecell}
\qquad $\Delta$ (vs Base) 
& \color{posgreen}{+0.126} & \color{posgreen}{+0.011} & \color{posgreen}{+0.007} & \color{negred}{-0.015} & \color{posgreen}{+0.160} & \color{posgreen}{+0.061} & \color{posgreen}{+0.012} & \color{posgreen}{+0.006} & \color{posgreen}{+0.111} & \color{posgreen}{+15.22\%} & \color{posgreen}{+1.08\%} \\
\bottomrule
\end{tabular}
}
\vspace{-5pt}
\label{tab:main_results}
\end{table*}

\vspace{-5pt}
\begin{figure}[!t]
    \centering
    \includegraphics[width=0.95\linewidth]{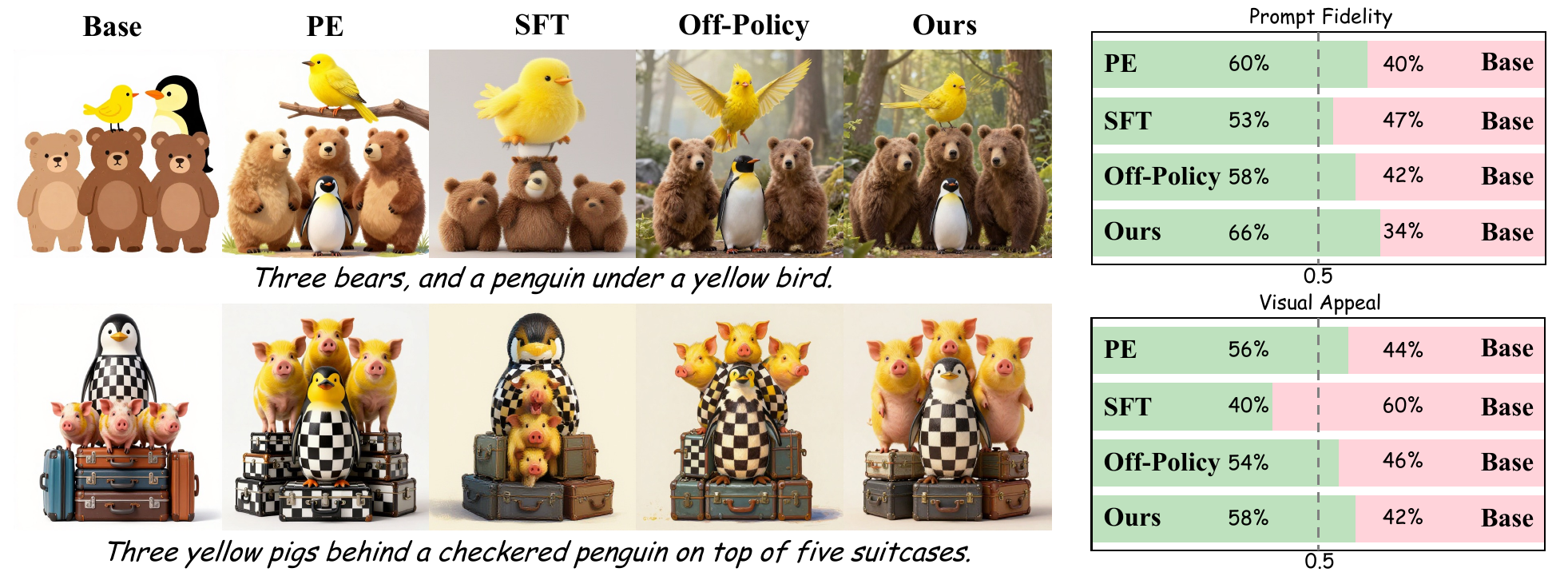}
    \vspace{-5pt}
    \caption{\textbf{Qualitative comparison and human preference study on Z-Image-Turbo.} We use \texttt{GPT-5.6 Sol} as PE. \textbf{Left:} visual comparisons among Base, PE, SFT, Off-policy distillation, and our method. \textbf{Right:} pairwise human preferences against Base in prompt fidelity and visual appeal.}
    \label{fig:vis}
    \vspace{-8pt}
\end{figure}

\paragraph{Applicability to post-trained models.}
PE-OPSD remains effective on the already post-trained Z-Image-Turbo (post-trained by Decoupled DMD~\citep{liu2026decoupled} and DMDR~\citep{jiang2026dmdr}). SFT yields smaller fidelity gains and a 1.40\% decline in visual appeal, whereas PE-OPSD improves both. These results support its applicability to models that have already undergone post-training.

\paragraph{Qualitative results and human preference study.}
Figure~\ref{fig:vis} presents visual examples and human preference study results. In pairwise comparisons against Base, human raters prefer PE-OPSD in 66\% of comparisons for prompt fidelity and 58\% for visual appeal. These are the highest observed preference rates among the evaluated methods, complementing the quantitative results. Human study details are provided in Appendix~\ref{app:human}.

\begin{figure*}[t]
    \centering
    \begin{minipage}[t]{0.6\textwidth}
        \vspace{0pt}
        \paragraph{Efficiency.}
        Table~\ref{tab:eff} compares prompt fidelity and latency on Z-Image. Under both evaluated PEs, PE-OPSD improves all three fidelity metrics over inference-time PE while retaining the measured latency of Base. It is $1.96\times$ and $4.51\times$ faster than deploying \texttt{PromptEnhancer-7B} and \texttt{PromptEnhancer-32B}, respectively. Thus, PE-OPSD transfers the benefits of prompt enhancement to the generator without requiring a PE for each inference request. More results about training costs are reported in Appendix~\ref{app:effcost}.
    \end{minipage}
    \hfill
    \begin{minipage}[t]{0.38\textwidth}
        \vspace{0pt}
        \centering
        \captionof{table}{\textbf{Performance and latency.}}
        \vspace{-10pt}
        \label{tab:eff}
        \tiny
        \setlength{\tabcolsep}{1.3mm}
        \renewcommand{\arraystretch}{0.9}
        \begin{tabular}{lcccc}
            \toprule
            \textbf{Method}
            & \textbf{GE}
            & \textbf{GE2$_{\texttt{GM}}$}
            & \textbf{GE2$_{\texttt{AM}}$}
            & \textbf{Lat. (s)} \\
            \midrule
            \rowcolor{lightgraycell}
            \multicolumn{5}{c}{\textbf{\texttt{PromptEnhancer-7B as PE}}} \\
            % \midrule
            Base
            & 0.650 & 0.306 & 0.761 & 9.69 \\
            Base+PE
            & 0.684 & 0.375 & 0.801 & 19.00 \\
            \rowcolor{lightbluecell}
            Base+Ours
            & 0.782 & 0.422 & 0.819 & 9.69 \\
            \rowcolor{lightbluecell}
            \quad $\Delta$ (vs +PE)
            & \color{posgreen}{+0.098} & \color{posgreen}{+0.047} & \color{posgreen}{+0.018} & \color{posgreen}{1.96$\times$} \\
            \midrule
            \rowcolor{lightgraycell}
            \multicolumn{5}{c}{\textbf{\texttt{PromptEnhancer-32B as PE}}} \\
            % \midrule
            Base
            & 0.650 & 0.306 & 0.761 & 9.69 \\
            Base+PE
            & 0.815 & 0.470 & 0.844 & 43.74 \\
            \rowcolor{lightbluecell}
            Base+Ours
            & 0.868 & 0.476 & 0.846 & 9.69 \\
            \rowcolor{lightbluecell}
            \quad $\Delta$ (vs +PE)
            & \color{posgreen}{+0.053} & \color{posgreen}{+0.006} & \color{posgreen}{+0.002} & \color{posgreen}{4.51$\times$} \\
            \bottomrule
        \end{tabular}
    \end{minipage}
\vspace{-5pt}
\end{figure*}

\begin{table*}[t]
\centering
\setlength{\tabcolsep}{1.25mm}
\caption{\textbf{Result across different PEs.} \textbf{Bold:} best; \underline{Underlined:} second-best.}
\vspace{-10pt}
{
\scriptsize
\renewcommand{\arraystretch}{0.95}
\begin{tabular}{lccccccccccc}
\toprule
\multirow{2}{*}[-0.8ex]{\textbf{Method}}
& \multicolumn{4}{c}{\textbf{GenEval (GE) Task}}
& \multicolumn{5}{c}{\textbf{GenEval2 (GE2) Task}} \\
\cmidrule(lr){2-5}
\cmidrule(lr){6-10}
& \textbf{GE}
& \textbf{PickScore}
& \textbf{CLIP}
& \textbf{Aes.}
& \textbf{GE2$_{\texttt{GM}}$}
& \textbf{GE2$_{\texttt{AM}}$}
& \textbf{PickScore}
& \textbf{CLIP}
& \textbf{Aes.} 
& \multirow{2}{*}[+3.0ex]{$\mathbb{I}_{\mathrm{PF}}$}
& \multirow{2}{*}[+3.0ex]{$\mathbb{I}_{\mathrm{VA}}$}
\\
\midrule
\rowcolor{lightgraycell}
\multicolumn{12}{c}{\textbf{\texttt{GPT-5.6 Sol as PE}}} \\
% \midrule
Z-Image
& 0.650 & 0.875 & 0.288 & 5.288 & 0.306 & 0.761 & 0.871 & 0.319 & 5.421 & 0.00\% & 0.00\%  \\
Z-Image+PE
& 0.810 & 0.899 & 0.299 & \textbf{5.433} & \underline{0.404} & 0.827 & 0.893 & 0.326 & \underline{5.638} & 14.27\% & 3.00\%\\
SFT
& 0.821 & 0.893 & \textbf{0.302} & 5.426 & 0.403 & \textbf{0.834} & 0.887 & 0.326 & 5.466 & \underline{14.93\%} & 1.83\% \\
Off-Policy Distillation
& \underline{0.824} & \underline{0.901} & \underline{0.300} & \underline{5.432} & 0.395 & 0.828 & \underline{0.897} & \underline{0.327} & 5.629 & 14.27\% & \underline{3.12\%} \\
\rowcolor{lightbluecell}
PE-OPSD (Ours)
& \textbf{0.846} & \textbf{0.902} & \textbf{0.302} & 5.423 & \textbf{0.406} & \underline{0.831} & \textbf{0.900} & \textbf{0.330} & \textbf{5.650} & \textbf{16.07\%} & \textbf{3.30\%} \\
\rowcolor{lightbluecell}
\qquad $\Delta$ (vs Base) 
& \color{posgreen}{+0.196} & \color{posgreen}{+0.027} & \color{posgreen}{+0.014} & \color{posgreen}{+0.135} & \color{posgreen}{+0.100} & \color{posgreen}{+0.070} & \color{posgreen}{+0.029} & \color{posgreen}{+0.011} & \color{posgreen}{+0.229} & \color{posgreen}{+16.07\%} & \color{posgreen}{+3.30\%}\\
\midrule
\rowcolor{lightgraycell}
\multicolumn{12}{c}{\textbf{\texttt{PromptEnhancer-7B as PE}}} \\
% \midrule
Z-Image
& 0.650 & 0.875 & 0.288 & 5.288 & 0.306 & 0.761 & 0.871 & 0.319 & 5.421 & 0.00\% & 0.00\% \\
Z-Image+PE
& 0.684 & 0.884 & 0.287 & \textbf{5.464} & 0.375 & 0.801 & 0.879 & 0.318 & \underline{5.589} & 6.28\% & 2.09\%\\
SFT
& 0.763 & 0.887 & \textbf{0.299} & \underline{5.424} & \underline{0.400} & \textbf{0.819} & 0.881 & \underline{0.325} & 5.424 & \underline{12.29\%} & 1.29\%\\
Off-Policy Distillation
& \underline{0.767} & \underline{0.896} & \underline{0.296} & 5.410 & 0.380 & \underline{0.804} & \underline{0.889} & 0.324 & 5.573 & 10.44\% & \underline{2.39\%}\\
\rowcolor{lightbluecell}
PE-OPSD (Ours)
& \textbf{0.782} & \textbf{0.898} & \underline{0.296} & 5.417 & \textbf{0.422} & \textbf{0.819} & \textbf{0.892} & \textbf{0.327} & \textbf{5.606} & \textbf{14.22\%} & \textbf{2.72\%}\\
\rowcolor{lightbluecell}
\qquad $\Delta$ (vs Base) 
& \color{posgreen}{+0.132} & \color{posgreen}{+0.023} & \color{posgreen}{+0.008} & \color{posgreen}{+0.129} & \color{posgreen}{+0.116} & \color{posgreen}{+0.058} & \color{posgreen}{+0.021} & \color{posgreen}{+0.008} & \color{posgreen}{+0.185} & \color{posgreen}{+14.22\%} & \color{posgreen}{+2.72\%}\\
\midrule
\rowcolor{lightgraycell}
\multicolumn{12}{c}{\textbf{\texttt{PromptEnhancer-32B as PE}}} \\
% \midrule
Z-Image
& 0.650 & 0.875 & 0.288 & 5.288 & 0.306 & 0.761 & 0.871 & 0.319 & 5.421 & 0.00\% & 0.00\% \\
Z-Image+PE
& 0.815 & 0.889 & 0.294 & \textbf{5.381} & \underline{0.470} & 0.844 & 0.879 & 0.325 & \textbf{5.515} & 18.77\% & \underline{1.50\%}\\
SFT
& 0.829 & 0.889 & \textbf{0.302} & \textbf{5.381} & 0.440 & \underline{0.845} & 0.877 & 0.329 & 5.373 & 18.07\% & 0.79\%\\
Off-Policy Distillation
& \underline{0.842} & \underline{0.895} & \underline{0.300} & \underline{5.363} & 0.459 & 0.843 & \underline{0.884} & \underline{0.331} & 5.455 & \underline{19.65\%} & 1.46\%\\
\rowcolor{lightbluecell}
PE-OPSD (Ours)
& \textbf{0.868} & \textbf{0.897} & \underline{0.300} & 5.352 & \textbf{0.476} & \textbf{0.846} & \textbf{0.888} & \textbf{0.334} & \underline{5.496} & \textbf{21.83\%} & \textbf{1.76\%}\\
\rowcolor{lightbluecell}
\qquad $\Delta$ (vs Base) 
& \color{posgreen}{+0.218} & \color{posgreen}{+0.022} & \color{posgreen}{+0.012} & \color{posgreen}{+0.064} & \color{posgreen}{+0.170} & \color{posgreen}{+0.085} & \color{posgreen}{+0.017} & \color{posgreen}{+0.015} & \color{posgreen}{+0.075} & \color{posgreen}{+21.83\%} & \color{posgreen}{+1.76\%}\\
\bottomrule
\end{tabular}
}
\label{tab:pe_results}
\vspace{-5pt}
\end{table*}

\vspace{-10pt}
\begin{table}[!t]
\centering
\caption{\textbf{Results on out-of-domain benchmarks.} For DPG-bench, we report Global (Glo.), Entity (Ent.), Attribute (Attr.), Relation (Rela.), Other and Overall$^\star$. For T2I-CompBench++, we report Color, Shape, Textual (Tex.), Numeracy (Num.), Complex (Comp.), Spatial (Spa.), 3D Spatial (3D Spa.), Non-spatial (Non-spa.) and Overall$^\star$. \textbf{Bold:} best in Overall$^\star$.}
\vspace{-7pt}
{
\tiny
\setlength{\tabcolsep}{1mm}
\renewcommand{\arraystretch}{0.9}
\begin{tabular}{lccccccccccccccccc}
\toprule
\multirow{2}{*}[-0.8ex]{\textbf{Method}}
& \multicolumn{6}{c}{\textbf{DPG-bench}}
& \multicolumn{9}{c}{\textbf{T2I-CompBench++}}
& \multicolumn{1}{c}{\textbf{EvalMuse}}
\\
\cmidrule(lr){2-7}
\cmidrule(lr){8-16}
\cmidrule(lr){17-17}
& \textbf{Overall$^\star$}
& \textbf{Glo.}
& \textbf{Ent.}
& \textbf{Attr.}
& \textbf{Rela.}
& \textbf{Other}
& \textbf{Overall$^\star$}
& \textbf{Color}
& \textbf{Shape}
& \textbf{Tex.}
& \textbf{Num.}
& \textbf{Comp.}
& \textbf{Spa.}
& \textbf{3D Spa.}
& \textbf{Non-spa.}
& \textbf{Overall$^\star$}
\\
\midrule
SD3.5-M
& \textbf{83.9} & 84.6 & 89.6 & 88.1 & 93.0 & 80.9 & 0.51 & 0.80 & 0.54 & 0.74 & 0.59 & 0.37 & 0.32 & 0.36 & 0.31 & 3.199\\
\rowcolor{lightbluecell}
\qquad +Ours
& 83.8 & 83.9 & 89.3 & 88.4 & 93.0 & 82.1 & \textbf{0.55} & 0.83 & 0.60 & 0.73 & 0.63 & 0.39 & 0.46 & 0.42 & 0.31 & \textbf{3.368}\\
\midrule
Z-Image
& 86.3 & 83.5 & 91.7 & 90.1 & 94.6 & 87.8 & 0.53 & 0.85 & 0.59 & 0.79 & 0.63 & 0.40 & 0.31 & 0.37 & 0.31 & 3.340\\
\rowcolor{lightbluecell}
\qquad +Ours
& \textbf{87.0} & 82.6 & 92.0 & 90.2 & 94.5 & 89.4 & \textbf{0.58} & 0.87 & 0.61 & 0.81 & 0.71 & 0.42 & 0.36 & 0.41 & 0.32 & \textbf{3.561}\\
\midrule
Z-Image-Turbo
& 84.5 & 78.6 & 91.2 & 88.3 & 93.2 & 88.4 & 0.53 & 0.81 & 0.56 & 0.75 & 0.69 & 0.40 & 0.36 & 0.41 & 0.31 & 3.515\\
\rowcolor{lightbluecell}
\qquad +Ours
& \textbf{85.0} & 77.8 & 91.2 & 88.9 & 93.9 & 88.1 & \textbf{0.56} & 0.89 & 0.53 & 0.76 & 0.72 & 0.40 & 0.52 & 0.42 & 0.31 & \textbf{3.534} \\
\bottomrule
\end{tabular}
}
\vspace{-8pt}
\label{tab:benchmark}
\end{table}

\subsection{Generalization}

\paragraph{Different PEs.}
On Z-Image, PE-OPSD achieves the highest $\mathbb{I}_{\mathrm{PF}}$ and $\mathbb{I}_{\mathrm{VA}}$ with each of the three tested PEs (Table~\ref{tab:pe_results}), demonstrating that its benefits extend beyond the default PE. The largest aggregate fidelity gain is obtained with \texttt{PromptEnhancer-32B}, while \texttt{GPT-5.6 Sol} yields the largest visual appeal gain, suggesting that PE choice affects the balance between these objectives.

\paragraph{Out-of-domain generalization.}
Table~\ref{tab:benchmark} shows that all three backbones improve their Overall scores on T2I-CompBench++ and EvalMuse, with consistent gains in spatial relations on T2I-CompBench++. On DPG-Bench, which features long and detailed prompts, PE-OPSD largely preserves the base models' performance. Overall, PE-OPSD generalizes across OOD benchmarks, improving compositional fidelity while preserving performance on already detailed prompts.

% \paragraph{Scalability.}
% The improvements extend to larger models such as FLUX.2-klein variants and QwenImage-2512 (Table~\ref{tab:scalingup}). PE-OPSD exceeds Base+PE in GE and GE2$_{\texttt{GM}}$ on all three configurations, with QwenImage-2512 achieving a 43.8\% aggregate fidelity improvement over Base. Aggregate visual appeal also improves over Base. Overall, these results demonstrate that PE-OPSD scales effectively to larger models, delivering stronger prompt fidelity while preserving visual appeal.
\paragraph{Scalability.}
We further evaluate PE-OPSD on the larger 9B FLUX.2-klein variants~\citep{blackforestlabs2026flux2klein} and the 20B QwenImage-2512~\citep{zhao2026qwenimage}, as summarized in Figure~\ref{fig:scaling_radar} (complete results are in Appendix~\ref{app:large}). Across three models, PE-OPSD improves upon the base model and inference-time PE on aggregate prompt fidelity. It also remains competitive across aggregate visual appeal over Base. These results demonstrate that PE-OPSD scales effectively to larger models.

\begin{figure}[!t]
    \centering
    \includegraphics[width=0.9\linewidth]{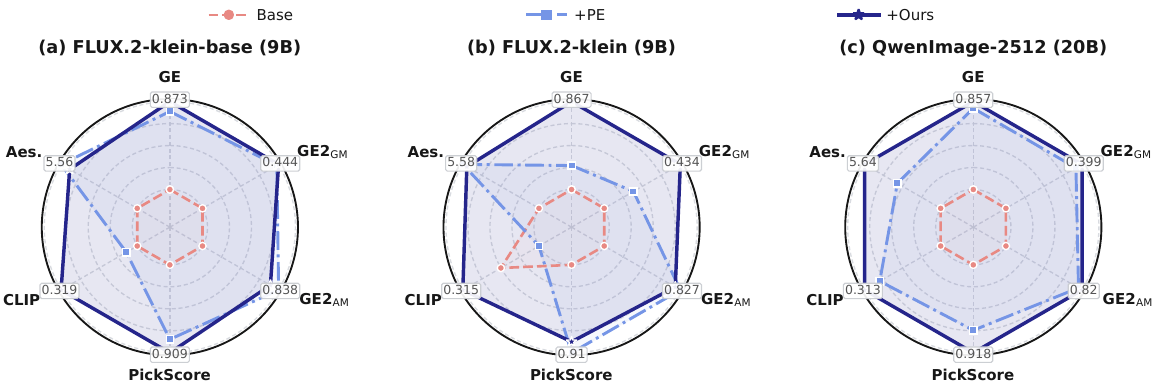}
    \vspace{-5pt}
    \caption{\textbf{Scaling to larger models.} PickScore, CLIP, and Aesthetics are averaged over GenEval and GenEval2. For visualization, each metric is independently normalized to $[0.3,1.0]$.}
    \label{fig:scaling_radar}
\end{figure}

\begin{figure}[!t]
    \centering
    \includegraphics[width=0.95\linewidth]{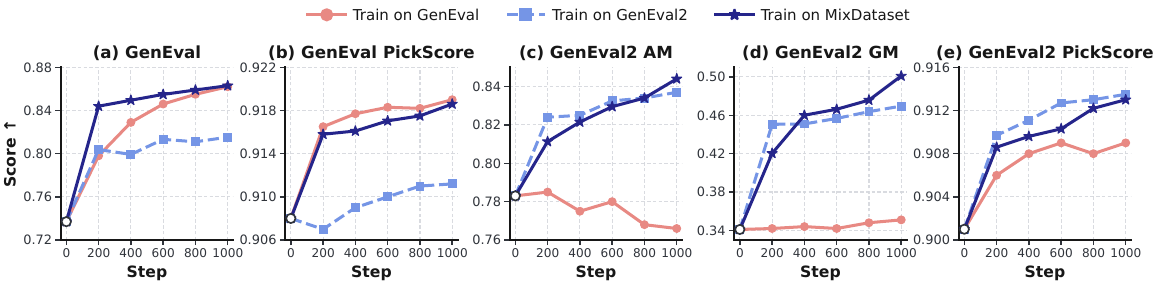}
    \vspace{-7pt}
    \caption{\textbf{Ablation study results on different training datasets.} We use MixDataset by default.}
    \label{fig:data}
\end{figure}

\begin{table}[!t]
% \vspace{-5pt}
\centering
\begin{minipage}{0.29\textwidth}
\centering
\caption{Ablation study results on different loss functions, which differ only in their timestep weighting.}
\label{tab:loss}
\vspace{-7pt}
{
\scriptsize
\setlength{\tabcolsep}{2mm}
\renewcommand{\arraystretch}{0.95}
\begin{tabular}{lccc}
\toprule
\textbf{Loss}
& \textbf{GE}
& \textbf{GE2$_{\texttt{GM}}$}
& \textbf{GE2$_{\texttt{AM}}$}\\
\midrule
$x_0$ & 0.863 & 0.474 & 0.834\\
$v$ & 0.866 & 0.487 & 0.838\\
\rowcolor{lightbluecell}
$\mu$ & 0.863 & 0.501 & 0.844\\
\bottomrule
\end{tabular}
}
\end{minipage}
\hfill
\begin{minipage}{0.36\textwidth}
\centering
\caption{Ablation study results on CFG settings. $^\dagger$denotes the unconditional branch is detached. Time (s): training times for one step.}
\label{tab:cfg}
\vspace{-7pt}
{
\scriptsize
\setlength{\tabcolsep}{1.4mm}
\renewcommand{\arraystretch}{0.9}
\begin{tabular}{lcccc}
\toprule
\textbf{Settings}
& \textbf{GE}
& \textbf{GE2$_{\texttt{GM}}$}
& \textbf{GE2$_{\texttt{AM}}$}
& \textbf{Time} \\
\midrule
CFG=4.0 & 0.834 & 0.430 & 0.835 & 99.13 \\
CFG=4.0$^\dagger$ & 0.835 & 0.421 & 0.834  & 73.39 \\
\rowcolor{lightbluecell}
w/o CFG & 0.846 & 0.406 & 0.831 & 49.15 \\
\bottomrule
\end{tabular}
}
\end{minipage}
\hfill
\begin{minipage}{0.3\textwidth}
\centering
\caption{Ablation study results on the steps for training. Time (s): training times for one step.}
\label{tab:timestep}
\vspace{-7pt}
{
\scriptsize
\setlength{\tabcolsep}{1.2mm}
\renewcommand{\arraystretch}{0.95}
\begin{tabular}{lcccc}
\toprule
\textbf{Step}
& \textbf{GE}
& \textbf{GE2$_{\texttt{GM}}$}
& \textbf{GE2$_{\texttt{AM}}$}
& \textbf{Time} \\
\midrule
2 & 0.862 & 0.469 & 0.835 & 7.47 \\
\rowcolor{lightbluecell}
4 & 0.863 & 0.501 & 0.844 & 14.31 \\
8 & 0.863 & 0.483 & 0.837 & 28.16 \\
\bottomrule
\end{tabular}
}
\end{minipage}
\vspace{-10pt}
\end{table}

% \begin{figure}[!t]
%     \centering
%     \vspace{-10pt}
%     \includegraphics[width=0.93\linewidth]{data.pdf}
%     \vspace{-5pt}
%     \caption{Ablation study results on different training datasets. We use MixDataset by default.}
%     \label{fig:data}
%     \vspace{-10pt}
% \end{figure}

\subsection{Ablation Studies}

\paragraph{Loss.}
Among losses differing only in timestep weighting, $\mu$-loss achieves the highest GE2 scores while remaining close to $v$-loss on GE (Table~\ref{tab:loss}), motivating its use as the default.
\vspace{-3pt}
\paragraph{CFG.}
Omitting CFG roughly halves training time and improves GE, but lowers GE2 scores (Table~\ref{tab:cfg}). Detaching the unconditional branch partially reduces the computational overhead. We omit CFG by default for training efficiency.
\vspace{-3pt}
\paragraph{On-policy training step.}
We conduct this ablation study in \texttt{Z-Image-Turbo}. As shown in Table~\ref{tab:timestep}, 4 rollout steps achieve the highest GE2 scores at approximately half the iteration time of 8 steps, with nearly identical GE across the tested settings. We therefore use four steps by default.
\vspace{-3pt}
\paragraph{Training data.}
As shown in Figure~\ref{fig:data}, training on a single dataset leads to dataset-specific specialization. GenEval-only training performs well on GenEval but transfers poorly to GenEval2, whereas GenEval2-only training compromises performance on GenEval. In contrast, MixDataset achieves the best balance across both benchmarks and is therefore used as our default training set.
\vspace{-3pt}
\section{Conclusion}
We introduced PE-OPSD, an on-policy self-distillation framework that treats enhanced prompts as privileged training information. PE-OPSD transfers the generation behavior induced by enhanced prompts to a raw-prompt generator through supervision on student-generated trajectories. Across three main backbones, it improves aggregate prompt fidelity by 13.35\%-16.07\% over Base and exceeds inference-time PE, SFT, and off-policy distillation while retaining base-model latency. Results with alternative PEs, larger models, and out-of-domain benchmarks further support its applicability.

\bibliography{iclr2027_conference}
\bibliographystyle{iclr2027_conference}

\clearpage
\appendix

\section{From Trajectory KL to Deterministic Matching}
\label{app:kl_to_ode}

The main text directly presents the deterministic objective used in training. Here, we provide a probabilistic interpretation through trajectory-level distillation and show how it motivates a deterministic matching surrogate on student-visited states.

\paragraph{Trajectory-level on-policy distillation.}
For a prompt pair $(p,p^{+})$, let the student and teacher trajectory distributions be
\begin{align}
    \Pi_{\theta}^{S}(\tau\mid p)
    &=
    p(x_{t_K})
    \prod_{k=1}^{K}
    \pi_{\theta,k}^{S}
    (x_{t_{k-1}}\mid x_{t_k},p),\\
    \Pi_{\bar{\theta}}^{T}(\tau\mid p^{+})
    &=
    p(x_{t_K})
    \prod_{k=1}^{K}
    \pi_{\bar{\theta},k}^{T}
    (x_{t_{k-1}}\mid x_{t_k},p^{+}),
\end{align}
where both distributions share the same initial noise prior. A trajectory-level distillation objective is
\begin{equation}
    \mathcal{L}_{\mathrm{traj}}
    =
    \mathbb{E}_{(p,p^{+})\sim\mathcal{D}}
    \left[
    D_{\mathrm{KL}}\!\left(
        \Pi_{\theta}^{S}(\cdot\mid p)
        \,\middle\|\,
        \Pi_{\bar{\theta}}^{T}(\cdot\mid p^{+})
    \right)
    \right].
    \label{eq:app_traj_kl}
\end{equation}
By the chain rule for KL divergence, this objective decomposes as
\begin{equation}
\begin{aligned}
    \mathcal{L}_{\mathrm{traj}}
    =
    \mathbb{E}_{(p,p^{+})\sim\mathcal{D}}
    \left[
    \sum_{k=1}^{K}
    \mathbb{E}_{x_{t_k}\sim d_{\theta,k}(\cdot\mid p)}
    \left[
    D_{\mathrm{KL}}\!\left(
        \pi_{\theta,k}^{S}
        \,\middle\|\,
        \pi_{\bar{\theta},k}^{T}
    \right)
    \right]
    \right],
\end{aligned}
\label{eq:app_kl_decomposition}
\end{equation}
where $d_{\theta,k}(\cdot\mid p)$ is the state distribution induced by the current raw-prompt student. Thus, the teacher is queried at states visited by the student rather than states from its own enhanced-prompt trajectory.

\paragraph{Conditional Gaussian transitions.}
At a fixed student-visited state $(x_{t_k},t_k)$, consider Gaussian transitions with shared covariance~\citep{liu2026flow,li2026diffusionopd,lin2026dreopd}:
\begin{equation}
    \pi_k^{b}
    =
    \mathcal{N}(\mu_k^{b},\Sigma_k),
    \qquad
    \mu_k^{b}
    =
    x_{t_k}-\Delta t_k v_k^{b},
    \qquad b\in\{S,T\},
    \label{eq:app_gaussian_transition}
\end{equation}
where $\Sigma_k\succ 0$ is shared by the student and teacher. Their conditional KL divergence is
\begin{equation}
    D_{\mathrm{KL}}\!\left(
        \pi_k^{S}\,\middle\|\,\pi_k^{T}
    \right)=
    \frac{1}{2}
    \left\|
        \mu_k^{S}-\mu_k^{T}
    \right\|_{\Sigma_k^{-1}}^{2}=
    \frac{(\Delta t_k)^2}{2}
    \left\|
        v_k^{S}-v_k^{T}
    \right\|_{\Sigma_k^{-1}}^{2}.
\label{eq:app_conditional_kl}
\end{equation}
For any positive-definite shared covariance, the unique pointwise minimizer is $v_k^{S}=v_k^{T}$, equivalently $\mu_k^{S}=\mu_k^{T}$.The covariance affects the weighting of the regression objective but not its pointwise optimum.

\paragraph{From stochastic transitions to deterministic matching.}
For two deterministic transitions with distinct endpoints, the corresponding Dirac measures are mutually singular, and their KL divergence is therefore infinite~\citep{gray2011entropy}. We thus do not obtain the deterministic objective by directly taking a zero-variance KL limit. Instead, we retain the pointwise optimizer of the shared-covariance Gaussian objective and realize it through deterministic $L2$-matching.
\begin{equation}
    \mathcal{L}_{\mu}
    =
    \mathbb{E}
    \left[
        \frac{1}{K}
        \sum_{k=1}^{K}
        \left\|
            \mu_k^{S}
            -
            \operatorname{sg}[\mu_k^{T}]
        \right\|_2^2
    \right]=
    \mathbb{E}
    \left[
        \frac{1}{K}
        \sum_{k=1}^{K}
        (\Delta t_k)^2
        \left\|
            v_k^{S}
            -
            \operatorname{sg}[v_k^{T}]
        \right\|_2^2
    \right].
\label{eq:app_deterministic_mu}
\end{equation}
The rollout states and teacher predictions are detached during each
optimization step. This deterministic objective preserves the
pointwise teacher-matching target of the Gaussian formulation while
avoiding the introduction of transition noise during training.

\paragraph{Alternative deterministic parameterizations.}
Besides matching Euler transition targets, we consider direct
velocity matching and clean-latent matching. Under the linear flow
interpolation
\begin{equation}
    x_t=(1-t)x_0+t\epsilon,
    \qquad
    v=\epsilon-x_0,
\end{equation}
the predicted clean latent is
\begin{equation}
    \hat{x}_{0,k}^{b}
    =
    x_{t_k}-t_kv_k^{b},
    \qquad b\in\{S,T\}.
\end{equation}
Consequently,
\begin{equation}
\begin{aligned}
    \left\|
        \mu_k^{S}-\operatorname{sg}[\mu_k^{T}]
    \right\|_2^2
    &=
    (\Delta t_k)^2
    \left\|
        v_k^{S}-\operatorname{sg}[v_k^{T}]
    \right\|_2^2,\\
    \left\|
        \hat{x}_{0,k}^{S}
        -\operatorname{sg}[\hat{x}_{0,k}^{T}]
    \right\|_2^2
    &=
    t_k^2
    \left\|
        v_k^{S}-\operatorname{sg}[v_k^{T}]
    \right\|_2^2.
\end{aligned}
\label{eq:app_parameterization_equivalence}
\end{equation}
The three deterministic objectives can therefore be written as
\begin{equation}
    \mathcal{L}
    =
    \mathbb{E}
    \left[
        \frac{1}{K}
        \sum_{k=1}^{K}
        \omega(t_k)
        \left\|
            v_k^{S}
            -
            \operatorname{sg}[v_k^{T}]
        \right\|_2^2
    \right],
\end{equation}
with
\begin{equation}
    \omega_v(t_k)=1,
    \qquad
    \omega_{\mu}(t_k)=(\Delta t_k)^2,
    \qquad
    \omega_{x_0}(t_k)=t_k^2.
\end{equation}
All three objectives share the pointwise optimum $v_k^{S}=v_k^{T}$, but they assign different weights to denoising timesteps and therefore need not produce identical optimization dynamics.

\section{Experimental Details}\label{app:exp}
\subsection{Training Data}
For GenEval~\citep{ghosh2023geneval}, we adopt the data splits released by Flow-GRPO~\citep{liu2026flow}, consisting of 50K training prompts and 2,212 evaluation instances. For GenEval2~\citep{kamath2025geneval2}, we use the officially released 20K synthetic prompts for training and the official set of 800 prompts for evaluation. Unless otherwise specified, all trainable methods use \emph{MixDataset}, formed by combining the GenEval and GenEval2 training sets. Evaluation prompts are never used for training.

We construct the prompt-pair dataset $\mathcal{D}=\{(p,p^{+})\}$ offline by applying each PE to the raw training prompts using the system prompt in Appendix~\ref{app:sp}. Since \texttt{GPT-5.6 Sol} is a general-purpose model rather than a PE specifically trained for prompt rewriting, its outputs may occasionally alter or omit subject identities or attributes. We therefore apply an automatic consistency check using the system prompt in Appendix~\ref{app:sp1}. Here, we use \texttt{GPT-5.6 Sol} as judge model. Rejected prompts are regenerated until they pass this check. This preprocessing is performed once before training, so the PE need not be loaded during optimization.

\subsection{Training Configurations.}
All experiments are conducted on one to four nodes with 8 NVIDIA A100 GPUs. We use a unified training recipe across model families and trainable baselines. Within each backbone, all methods use the same initialization, training data, and hyper-parameters.

To improve training efficiency, we collect student rollouts using fewer denoising steps than at inference, following the denoising-reduction practice established in prior work~\citep{liu2026flow,ping2026flow}. Specifically, we use 4 training rollout steps for Z-Image-Turbo and FLUX.2-klein, 10 for SD3.5-M, and 14 for Z-Image, FLUX.2-klein-base, and QwenImage-2512. Here, \emph{on-policy} refers to the provenance of the training states: they are generated by the current student under raw prompts. At every visited training state, the student and teacher are evaluated at the same latent and timestep. This does not require the training rollout and inference process to use identical timestep discretizations.

For evaluation, we set the number of inference steps to 4 for FLUX.2-klein, 8 for Z-Image-Turbo, 40 for SD3.5M, and 50 for Z-Image, FLUX.2-klein-base and QwenImage-2512 following the official settings. For evaluation CFG settings, we set 1.0 (disabled) for Z-Image-Turbo and FLUX.2-klein, 4.0 for Z-Image, FLUX.2-klein-base and QwenImage-2512, and 4.5 for SD3.5M.The other hyperparameter settings are reported in Table~\ref{tab:training_hyperparameters}. 

\begin{table*}[t]
\centering
\caption{\textbf{Default training hyperparameters for PE-OPSD.}
All models follow the same recipe unless otherwise specified.}
\label{tab:training_hyperparameters}
\setlength{\tabcolsep}{6mm}
\begin{tabular}{ll}
\toprule
\textbf{Configuration} & \textbf{Setting} \\
\midrule
Optimizer & AdamW \\
Learning rate & $1\times10^{-4}$, constant without warm-up \\
Optimizer momentum & $(\beta_1,\beta_2)=(0.9,0.999)$ \\
Adam $\epsilon$ / weight decay & $10^{-8}$ / $0$ \\
Gradient clipping & $1.0$ \\
Global batch size & $64$\\
Optimization steps & $1{,}000$ \\
LoRA rank / scaling factor & $64$ / $128$ \\
EMA decay $\gamma$ & $0.999$ \\
\midrule
Training resolution & All $512\times512$ \\
Training CFG & Disabled \\
Evaluation resolution & All $512\times512$ \\
Maximum text sequence length & $512$ \\
Distributed optimization & DeepSpeed ZeRO-2 \\
\bottomrule
\end{tabular}
\end{table*}

The ablation studies for different loss, dataset, and training step are conducted on Z-Image-Turbo using \texttt{GPT-5.6 Sol} as PE. The CFG ablation is conducted on Z-Image with the same PE. Unless explicitly varied, all remaining settings follow the default configuration.

\subsection{Models}
Table~\ref{tab:model_checkpoints} lists the checkpoints used in our experiments. 

\begin{table*}[t]
\centering
\caption{\textbf{Models and checkpoints used in our experiments.}}
\label{tab:model_checkpoints}

\setlength{\tabcolsep}{3mm}
\begin{tabular}{ll}
\toprule
\textbf{Model} & \textbf{Checkpoint} \\
\midrule
SD3.5-Medium & \href{https://huggingface.co/stabilityai/stable-diffusion-3.5-medium}{\texttt{stabilityai/stable-diffusion-3.5-medium}} \\
SD3.5-Large & \href{https://huggingface.co/stabilityai/stable-diffusion-3.5-large}{\texttt{stabilityai/stable-diffusion-3.5-large}} \\
Z-Image & \href{https://huggingface.co/Tongyi-MAI/Z-Image}{\texttt{Tongyi-MAI/Z-Image}} \\
Z-Image-Turbo & \href{https://huggingface.co/Tongyi-MAI/Z-Image-Turbo}{\texttt{Tongyi-MAI/Z-Image-Turbo}} \\
FLUX.1-dev & \href{https://huggingface.co/black-forest-labs/FLUX.1-dev}{\texttt{black-forest-labs/FLUX.1-dev}} \\
FLUX.2-klein-base-9B & \href{https://huggingface.co/black-forest-labs/FLUX.2-klein-base-9B}{\texttt{black-forest-labs/FLUX.2-klein-base-9B}} \\
FLUX.2-klein-9B & \href{https://huggingface.co/black-forest-labs/FLUX.2-klein-9B}{\texttt{black-forest-labs/FLUX.2-klein-9B}} \\
QwenImage-2512 & \href{https://huggingface.co/Qwen/Qwen-Image-2512}{\texttt{Qwen/Qwen-Image-2512}} \\
\midrule
PromptEnhancer-7B & \href{https://huggingface.co/tencent/HunyuanImage-2.1/tree/main/reprompt}{\texttt{tencent/HunyuanImage-2.1/reprompt}} \\
PromptEnhancer-32B & \href{https://huggingface.co/PromptEnhancer/PromptEnhancer-32B}{\texttt{PromptEnhancer/PromptEnhancer-32B}} \\
\bottomrule
\end{tabular}
\end{table*}

\subsection{Baselines}
\paragraph{Inference-time PE.}
\emph{Base} generates directly from the raw prompt $p$, whereas \emph{Base+PE} applies the PE and conditions the frozen generator on $p^{+}$ at inference time.

\paragraph{Supervised fine-tuning.}
For SFT, we first use each frozen base model to generate pseudo-target images from enhanced prompts under its corresponding inference configuration. We then fine-tune the model with the standard flow-matching objective on raw-prompt/image pairs $(p,x^{+})$. Thus, SFT transfers enhanced-prompt behavior through offline-generated images rather than online vector-field supervision.

\paragraph{Off-policy distillation.}
Off-policy distillation uses the same configuration as PE-OPSD. The only difference is the rollout distribution: off-policy states are generated by the teacher conditioned on $p^{+}$, whereas PE-OPSD evaluates the teacher on states visited by the student conditioned on $p$.

\subsection{Benchmarks}
\paragraph{In-domain evaluation.}
GenEval~\citep{ghosh2023geneval} evaluates object presence, counting, colors, spatial relations, and attribute binding using its detector-based protocol. GenEval2~\citep{kamath2025geneval2} contains 800 prompts with broader visual concepts and higher compositional complexity. We follow its official Soft-TIFA evaluation, reporting the geometric mean (GE2$_{\mathrm{GM}}$) for prompt-level correctness and the arithmetic mean (GE2$_{\mathrm{AM}}$) for atom-level correctness. On both benchmarks, we additionally report CLIP score~\cite{hessel2021clipscore}, PickScore~\citep{kirstain2023pick} normalized by 26, and aesthetics score~\citep{schuhmann2022aesthetics}.

\paragraph{Out-of-domain evaluation.}
DPG-Bench~\citep{hu2024dpg} evaluates dense prompt following over Global, Entity, Attribute, Relation, and Other categories. T2I-CompBench++~\citep{huang2025t2i} evaluates attribute binding, numeracy, spatial and non-spatial relations, and complex compositions. For EvalMuse~\citep{han2026evalmuse}, we follow the original protocol using its 200 representative prompts and the FGA-BLIP2 overall alignment score. All Overall scores are computed using the corresponding official evaluation procedures.

\section{Human Evaluation Protocol}
\label{app:human}

\paragraph{Evaluation data and comparisons.}
We conduct human evaluation on all 800 prompts from the GenEval2 test set using Z-Image-Turbo as the generator and \texttt{GPT-5.6 Sol} as the PE. For each prompt, we compare Base against Base+PE, SFT, off-policy distillation, and PE-OPSD.

\paragraph{Evaluation interface and aggregation.}
Each comparison presents two generated images side by side, with the assignment of systems to the left and right positions randomized independently for every pair. Model identities and method names are hidden from the assessors, while the corresponding raw prompt is displayed above the images. Assessors provide two independent judgments (Left better or Right better): (1) \emph{Prompt Fidelity}, indicating which image more faithfully satisfies the objects, attributes, counts, and relationships specified by the raw prompt; and (2) \emph{Visual Appeal}, indicating which image has better overall perceptual quality and aesthetics. For each criterion, the assessor selects either the left or right image. For the result aggregation, each image pair is independently evaluated by all three assessors for both criteria. We determine the preference for each pair by majority vote.

\paragraph{Assessors and quality control.}
We recruit three professional assessors who are formally contracted and compensated at locally competitive rates. Before participating in the study, all assessors are informed of the task and potential exposure to generated content. They receive detailed criterion-specific instructions and complete a qualification test covering representative evaluation cases.

\section{Additional Experimental Results}\label{app:result}

\subsection{Training Efficiency}
Table~\ref{tab:training_efficiency} reports the method-specific cost of post-training. Although SFT has a substantially lower per-step cost, it requires generating pseudo-target images for the entire training set. This preprocessing dominates its total cost for SD3.5-M and Z-Image, for which PE-OPSD reduces the training time from $7.0$ to $3.1$ hours and from $28.4$ to $13.6$ hours, respectively. For Z-Image-Turbo, few-step sampling makes pseudo-target generation inexpensive, and SFT is slightly faster overall. However, SFT yields lower performance than both off-policy distillation and PE-OPSD (see Table~\ref{tab:main_results}).

PE-OPSD and off-policy distillation have identical training costs because they use the same number of rollout and teacher evaluations, differing only in whether states are collected from the student or teacher trajectory. Overall, PE-OPSD avoids the additional image-generation and storage requirements of SFT while incurring the expected cost of online rollout supervision. Offline enhanced-prompt construction is shared by all trainable methods and is therefore excluded from this method-specific comparison.

\begin{table*}[h]
\centering
\caption{\textbf{Training efficiency in our experiments.} One-step denotes the wall-clock time per optimization step under the default configuration, and Full denotes the cost of 1,000 optimization steps. For SFT, Full$^\dagger$ is reported as training time + offline pseudo-target generation time.}
\vspace{-5pt}
\label{tab:training_efficiency}
\setlength{\tabcolsep}{3mm}
\small
\begin{tabular}{lcccccc}
\toprule
\multirow{2}{*}[-0.8ex]{\textbf{Model}}
& \multicolumn{2}{c}{\textbf{SFT}}
& \multicolumn{2}{c}{\textbf{Off-policy}} 
& \multicolumn{2}{c}{\textbf{PE-OPSD}} 
\\
\cmidrule(lr){2-3}
\cmidrule(lr){4-5}
\cmidrule(lr){6-7}
& \textbf{One-step}
& \textbf{Full$^\dagger$}
& \textbf{One-step}
& \textbf{Full}
& \textbf{One-step}
& \textbf{Full}
\\
\midrule
SD3.5-M & 1.47$s$ & 0.4$+$6.6$h$ & 11.02$s$ & 3.1$h$ & 11.02$s$ & 3.1$h$ \\
Z-Image & 3.14$s$ & 0.9$+$27.5$h$ & 49.15$s$ & 13.6$h$ & 49.15$s$ & 13.6$h$ \\
Z-Image-Turbo & 3.14$s$ & 0.9$+$2.5$h$ & 14.31$s$ & 4.0$h$ & 14.31$s$ & 4.0$h$ \\
\bottomrule
\end{tabular}
\vspace{-10pt}
\end{table*}

\subsection{Inference Latency}
\label{app:effcost}

Table~\ref{tab:latency} compares deployment latency with and without inference-time PE. Because PE-OPSD generates directly from raw prompts using the original inference pipeline, it retains the latency of Base across all evaluated backbones. In contrast, inference-time PE introduces a substantial fixed overhead, particularly for efficient few-step generators. Relative to \texttt{PromptEnhancer-7B}, PE-OPSD provides $1.52\times$--$12.21\times$ speedups; with \texttt{PromptEnhancer-32B}, the speedups increase to $3.31\times$--$50.67\times$. The largest gains occur on Z-Image-Turbo and FLUX.2-klein, where prompt rewriting is considerably more expensive than image generation itself. These results demonstrate that PE-OPSD preserves the benefits of enhanced prompts without adding per-request deployment latency.

\begin{table}[h]
\centering
\caption{\textbf{Inference latency and speedup across different settings.} Latency is the average generation time per image in seconds. +PE includes both prompt rewriting and image generation, whereas +Ours generates directly from the raw prompt without invoking PE.}
\vspace{-5pt}
\label{tab:latency}
\setlength{\tabcolsep}{5mm}
\small
\begin{tabular}{lcccc}
\toprule
\multirow{2}{*}[-0.8ex]{\textbf{Model}}
& \multicolumn{3}{c}{\textbf{Latency (s)}} \\
\cmidrule(lr){2-4}
& \textbf{Base}
& \textbf{+PE}
& \textbf{+Ours}
& \multirow{2}{*}[+3.0ex]{\textbf{Speedup}} 
\\
\midrule
\rowcolor{lightgraycell}
\multicolumn{5}{c}{\textbf{\texttt{PromptEnhancer-7B as PE}}} \\
SD3.5-M & 2.70 & 10.36 & 2.70 & \cellcolor{lightbluecell}{3.84$\times$}\\
Z-Image & 9.69 & 19.00 & 9.69 & \cellcolor{lightbluecell}{1.96$\times$}\\
Z-Image-Turbo & 0.88 & 8.53 & 0.88 & \cellcolor{lightbluecell}{9.69$\times$}\\
FLUX.2-klein & 0.67 & 8.18 & 0.67 & \cellcolor{lightbluecell}{12.21$\times$}\\
FLUX.2-klein-base & 14.45 & 21.97 & 14.45 & \cellcolor{lightbluecell}{1.52$\times$}\\
QwenImage-2512 & 13.38 & 22.09 & 13.38 & \cellcolor{lightbluecell}{1.65$\times$} \\
\midrule
\rowcolor{lightgraycell}
\multicolumn{5}{c}{\textbf{\texttt{PromptEnhancer-32B as PE}}} \\
SD3.5-M & 2.70 & 35.91 & 2.70 & \cellcolor{lightbluecell}{13.30$\times$}\\
Z-Image & 9.69 & 43.74 & 9.69 & \cellcolor{lightbluecell}{4.51$\times$}\\
Z-Image-Turbo & 0.88 & 34.21 & 0.88 & \cellcolor{lightbluecell}{38.88$\times$}\\
FLUX.2-klein & 0.67 & 33.95 & 0.67 & \cellcolor{lightbluecell}{50.67$\times$}\\
FLUX.2-klein-base & 14.45 & 47.77 & 14.45 & \cellcolor{lightbluecell}{3.31$\times$}\\
QwenImage-2512 & 13.38 & 47.33 & 13.38 & \cellcolor{lightbluecell}{3.54$\times$} \\
\bottomrule
\end{tabular}
\vspace{-5pt}
\end{table}

\subsection{Effect of EMA Teacher}
\begin{figure}[h]
    \centering
    \includegraphics[width=0.9\linewidth]{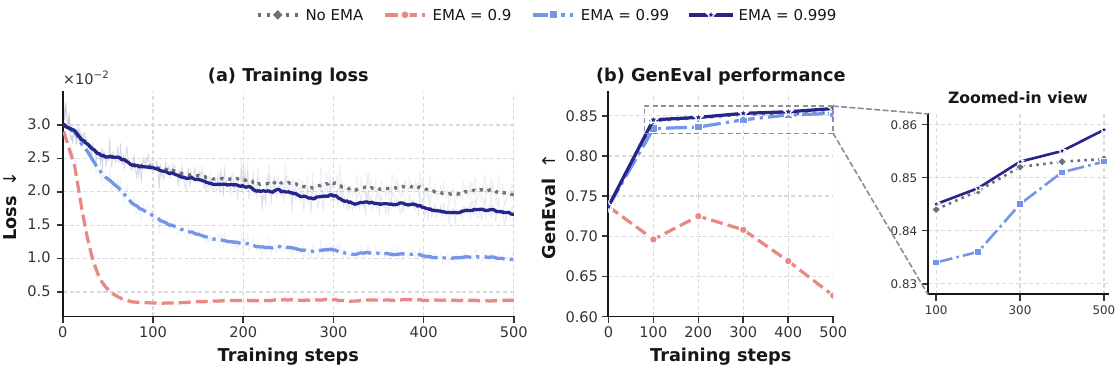}
    \vspace{-5pt}
    \caption{\textbf{Training dynamics and GenEval performance with different EMA settings.} Left: the loss curves where faint lines denote raw losses and bold lines show a the moving average; Right: the corresponding GenEval performance curves.}
    \label{fig:ema_loss}
\end{figure}

\begin{figure}[h]
    \centering
    \includegraphics[width=0.9\linewidth]{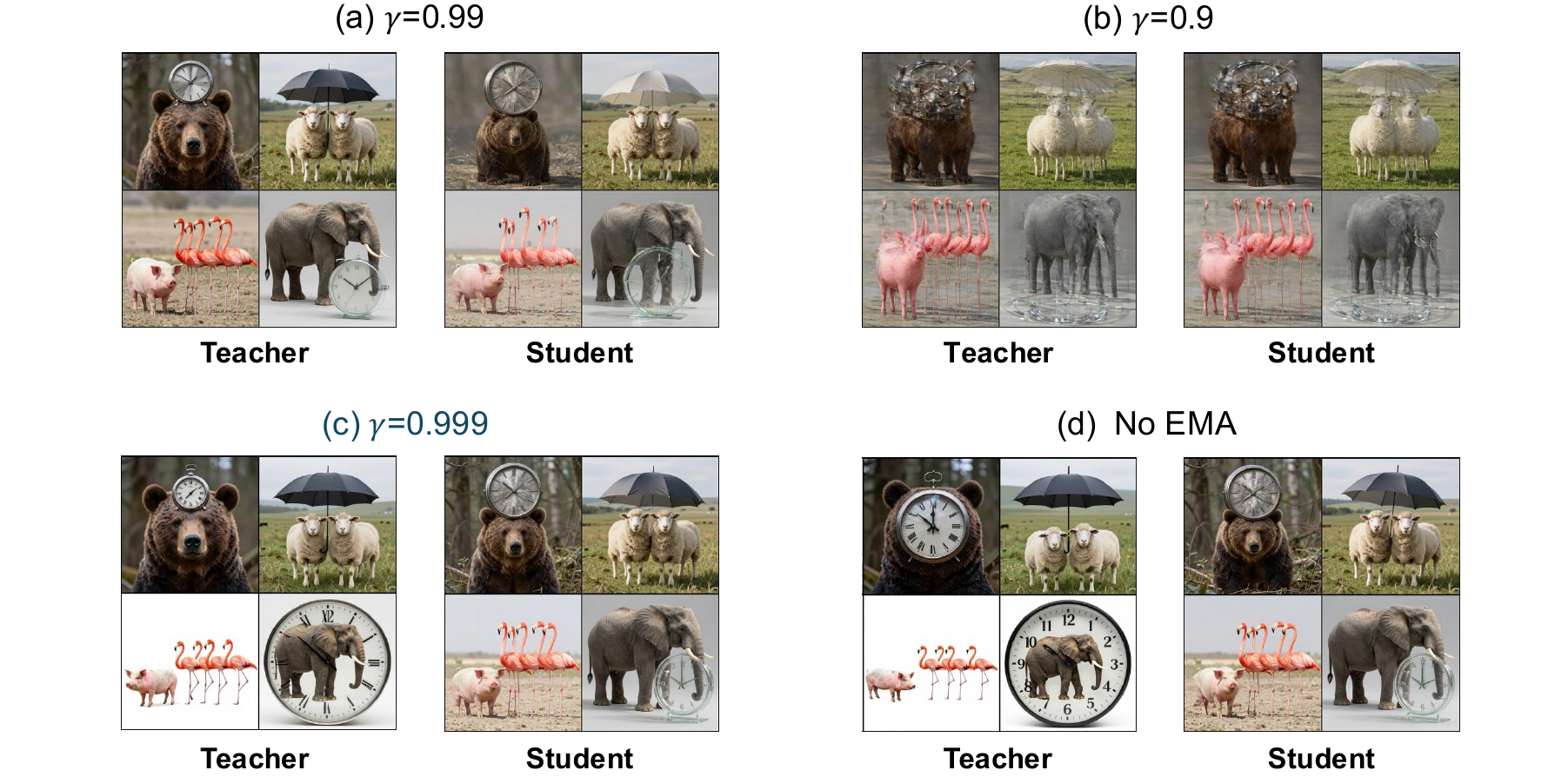}
    \vspace{-5pt}
    \caption{\textbf{Visual examples.} We visualize the samples generated by the student and teacher across different EMA settings at 500 training steps.}
    \label{fig:ema_vis}
    \vspace{-10pt}
\end{figure}

As shown in Figure~\ref{fig:ema_loss} and Figure~\ref{fig:ema_vis}, we study the effect of EMA teacher using four settings: a frozen Base teacher without EMA updates (No EMA), $\gamma=0.99$, $\gamma=0.9$, and our default $\gamma=0.999$. We conduct the experiments using Z-Image-Turbo with 500 training steps. 


The frozen teacher performs similarly to the default setting during early training but reaches a lower performance ceiling, indicating that allowing the teacher to evolve with the student provides stronger supervision at later stages. With $\gamma=0.99$, generated images remain visually coherent, although performance is slightly lower than with $\gamma=0.999$.

A more aggressive update with $\gamma=0.9$ substantially degrades generation quality. Although its training loss decreases rapidly, generated images contain pronounced artifacts and evaluation scores fall below Base. This suggests that an overly responsive teacher becomes too tightly coupled to the student and provides insufficiently stable targets. We therefore use $\gamma=0.999$, which balances teacher adaptation with temporal stability.

\subsection{Detailed Results of scalability}\label{app:large}

Table~\ref{tab:scalingup} reports the complete results on larger models. PE-OPSD achieves the highest aggregate prompt-fidelity improvement across all three backbones and consistently leads on GE and GE2$_{\texttt{GM}}$. It also improves aggregate visual appeal over Base, although Base+PE remains marginally better on the two FLUX variants. On QwenImage-2512, PE-OPSD obtains the best aggregate results for both objectives, further supporting its applicability to larger models.

\begin{table}[h]
\centering
\caption{\textbf{Results on scaling up to large models.} We use \texttt{GPT-5.6 Sol} as PE. PickScore is normalized by 26; Aes. denotes aesthetics; \textbf{Bold:} best; \underline{Underlined:} second-best.}
{
\scriptsize
\setlength{\tabcolsep}{1.25mm}
\renewcommand{\arraystretch}{1.0}
\begin{tabular}{lccccccccccc}
\toprule
\multirow{2}{*}[-0.8ex]{\textbf{Method}}
& \multicolumn{4}{c}{\textbf{GenEval (GE) Task}}
& \multicolumn{5}{c}{\textbf{GenEval2 (GE2) Task}} \\
\cmidrule(lr){2-5}
\cmidrule(lr){6-10}
& \textbf{GE}
& \textbf{PickScore}
& \textbf{CLIP}
& \textbf{Aes.}
& \textbf{GE2$_{\texttt{GM}}$}
& \textbf{GE2$_{\texttt{AM}}$}
& \textbf{PickScore}
& \textbf{CLIP}
& \textbf{Aes.} 
& \multirow{2}{*}[+3.0ex]{\textbf{$\mathbb{I}_{\mathrm{PF}}$}}
& \multirow{2}{*}[+3.0ex]{\textbf{$\mathbb{I}_{\mathrm{VA}}$}}
\\
\midrule
FLUX.2-klein-base (9B)
& 0.775 & 0.892 & \underline{0.303} & 5.140 & 0.359 & 0.773 & 0.880 & 0.328 & 5.310 & 0.00\% & 0.00\%\\
\qquad +PE
& \underline{0.862} & \underline{0.910} & 0.302 & \textbf{5.467} & \underline{0.442} & \textbf{0.838} & \underline{0.901} & \underline{0.330} & \textbf{5.658} & \underline{8.61\%} & \textbf{4.33\%}\\
\rowcolor{lightbluecell}
\qquad +Ours
& \textbf{0.873} & \textbf{0.913} & \textbf{0.305} & \underline{5.421} & \textbf{0.444} & \underline{0.831} & \textbf{0.905} & \textbf{0.333} & \underline{5.628} & \textbf{9.20\%} & \underline{4.16\%}\\
\rowcolor{lightbluecell}
\qquad $\Delta$ (vs Base) 
& \color{posgreen}{+0.098} & \color{posgreen}{+0.021} & \color{posgreen}{+0.002} & \color{posgreen}{+0.281} & \color{posgreen}{+0.085} & \color{posgreen}{+0.058} & \color{posgreen}{+0.025} & \color{posgreen}{+0.005} & \color{posgreen}{+0.318} & \color{posgreen}{+9.20\%} & \color{posgreen}{+4.16\%} \\
\midrule
FLUX.2-klein (9B)
& 0.856 & 0.911 & 0.299 & 5.288 & 0.348 & 0.797 & 0.901 & \textbf{0.329} & 5.430 & 0.00\% & 0.00\%\\
\qquad +PE
& \underline{0.859} & \textbf{0.915} & \underline{0.300} & \textbf{5.474} & \underline{0.381} & \textbf{0.827} & \underline{0.905} & 0.327 & \underline{5.682} & \underline{2.66\%} & \textbf{2.26\%}\\
\rowcolor{lightbluecell}
\qquad +Ours
& \textbf{0.867} & \underline{0.912} & \textbf{0.301} & \underline{5.433} & \textbf{0.434} & \underline{0.825} & \textbf{0.907} & \underline{0.328} & \textbf{5.700} & \textbf{5.98\%} & \underline{2.12\%}\\
\rowcolor{lightbluecell}
\qquad $\Delta$ (vs Base) 
& \color{posgreen}{+0.011} & \color{posgreen}{+0.001} & \color{posgreen}{+0.002} & \color{posgreen}{+0.145} & \color{posgreen}{+0.086} & \color{posgreen}{+0.028} & \color{posgreen}{+0.006} & \color{negred}{-0.001} & \color{posgreen}{+0.270} & \color{posgreen}{+5.98\%} & \color{posgreen}{+2.12\%} \\
\midrule
QwenImage-2512 (20B)
& 0.620 & 0.895 & 0.283 & 5.377 & 0.163 & 0.671 & 0.894 & 0.303 & \underline{5.822} & 0.00\% & 0.00\%\\
\qquad +PE
& \underline{0.839} & \underline{0.916} & \underline{0.297} & \textbf{5.462} & \underline{0.380} & \underline{0.813} & \underline{0.908} & \underline{0.321} & 5.782 & \underline{40.1\%} & \underline{1.20\%}\\
\rowcolor{lightbluecell}
\qquad +Ours
& \textbf{0.857} & \textbf{0.920} & \textbf{0.302} & \underline{5.421} & \textbf{0.399} & \textbf{0.820} & \textbf{0.916} & \textbf{0.324} & \textbf{5.857} & \textbf{43.8\%} & \textbf{1.67\%}\\
\rowcolor{lightbluecell}
\qquad $\Delta$ (vs Base) 
& \color{posgreen}{+0.237} & \color{posgreen}{+0.025} & \color{posgreen}{+0.019} & \color{posgreen}{+0.044} & \color{posgreen}{+0.236} & \color{posgreen}{+0.149} & \color{posgreen}{+0.022} & \color{posgreen}{+0.021} & \color{posgreen}{+0.035} & \color{posgreen}{+43.8\%} & \color{posgreen}{+1.67\%} \\
\bottomrule
\end{tabular}
}
\label{tab:scalingup}
\end{table}

\subsection{More Analysis in Table~\ref{tab:pe_results}}

\paragraph{Robustness across PEs.}
Table~\ref{tab:pe_results} shows that the performance of SFT, inference-time PE, and off-policy distillation varies with the chosen PE, whereas PE-OPSD consistently provides the strongest aggregate results. In particular, PE-OPSD improves $\mathbb{I}_{\mathrm{PF}}$ over off-policy distillation by $1.80$--$3.78$ percentage points and $\mathbb{I}_{\mathrm{VA}}$ by $0.18$--$0.33$ points across the three PEs. Since the two distillation methods share the teacher condition, objective, and optimization budget, this consistent margin supports the importance of supervising the student on states visited by its own raw-prompt trajectories.

\paragraph{Effect of the teacher condition.}
PE-OPSD also exceeds inference-time PE in both aggregate indices while requiring only raw prompts at deployment. Notably, its largest fidelity margin over Base+PE occurs with \texttt{PromptEnhancer-7B}, whose direct inference-time improvement is the weakest among the tested PEs. This indicates that the transferred benefit is not determined solely by the PE's one-shot inference performance. Meanwhile, \texttt{PromptEnhancer-32B} produces the strongest fidelity supervision, whereas \texttt{GPT-5.6 Sol} yields the largest visual-appeal improvement. Overall, PE-OPSD remains effective across PEs with substantially different capacities and enhancement behaviors.

\subsection{Different PEs}
\paragraph{Additional experimental results for different PEs.}

\begin{figure}[h]
    \centering
    \includegraphics[width=0.9\linewidth]{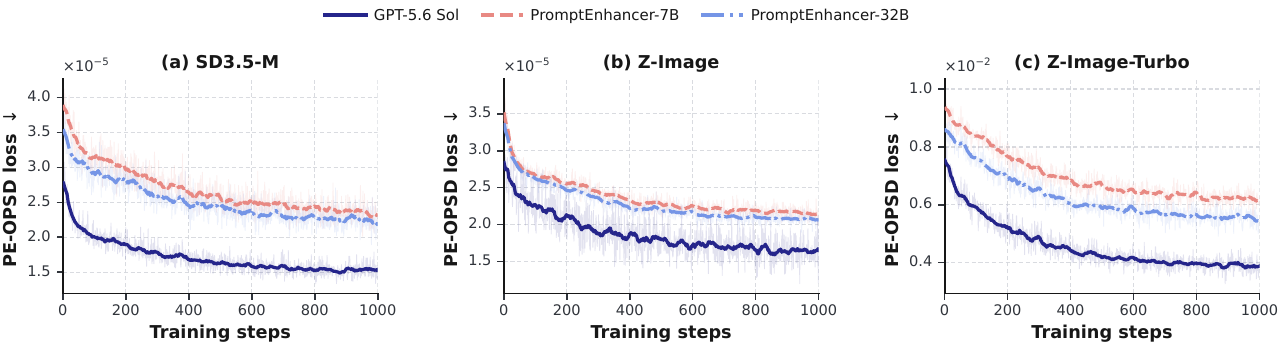}
    \vspace{-5pt}
    \caption{\textbf{Training dynamics with different PEs.} We report the PE-OPSD loss over 1,000 training steps. Faint lines denote raw losses, while bold lines show a the moving average.}
    \label{fig:pe_loss_curves}
\end{figure}

We further visualize the training dynamics with three PEs in our experiments across SD3.5-M, Z-Image, and Z-Image-Turbo. As shown in Figure~\ref{fig:pe_loss_curves}, the PE-OPSD loss consistently decreases and stabilizes within 1,000 training steps for all PEs and models. These similar optimization trends demonstrate that PE-OPSD remains stable across different PEs and does not rely on a specific PE.

\vspace{-5pt}
\vspace{-10pt}

\paragraph{Prompt examples.}
Figures~\ref{fig:prompt-comparison1}--\ref{fig:prompt-comparison3} show examples of raw prompts and their enhanced versions produced by different PEs.

\begin{figure*}[h]
\centering
\small
\begin{rawpromptbox}
five bears and a donut.
\end{rawpromptbox}
\begin{enhancedpromptbox}
Five bears gather in a loose semicircle around a single oversized glazed donut resting on the ground. Each bear has thick, textured brown fur, rounded ears, broad paws, and a distinct curious expression, with two bears seated at the front, two standing behind them, and the fifth leaning forward to inspect the pastry. The donut features a golden-brown surface, glossy pink icing, and scattered rainbow sprinkles, contrasting vividly with the bears' earthy coloring. Soft forest vegetation, moss-covered stones, and tall trees fill the background, while warm filtered sunlight creates gentle shadows and highlights the fur and shiny glaze. Whimsical, highly detailed storybook illustration with naturalistic anatomy and cinematic lighting.
\end{enhancedpromptbox}

\caption{\textbf{Comparison between the raw prompt and the enhanced prompt.} Here, we use \texttt{GPT-5.6 Sol} as PE.}
\label{fig:prompt-comparison1}
\end{figure*}

\begin{figure*}[h]
\centering
\small
\begin{rawpromptbox}
five bears and a donut.
\end{rawpromptbox}
\begin{enhancedpromptbox}
A group of five cartoon bears and one donut are presented in a cheerful arrangement against a plain, light-colored background. The bears are positioned in a line, with three standing and two sitting. The bear on the far left is a tall, slender figure with light brown fur, standing on its hind legs and waving a paw. Next to it, another tall bear with chocolate-brown fur stands with its head tilted curiously. In the center, a large, round brown bear sits comfortably, looking directly forward with a friendly expression. To its right, a shorter, honey-colored bear sits alongside the donut, looking up at it. On the far right, the fifth bear, a small grizzly-style bear with darker fur, sits with its paws on its knees. In front of the central sitting bear rests a large, classic ring donut, coated in a vibrant pink frosting and decorated with a scattering of multi-colored sprinkles. The image is rendered in a clean, digital illustration style with bold outlines and soft shading.
\end{enhancedpromptbox}
\caption{\textbf{Comparison between the raw prompt and the enhanced prompt.} Here, we use \texttt{PromptEnhancer-7B} as PE.}
\label{fig:prompt-comparison2}
\end{figure*}

\begin{figure*}[h]
\centering
\small
\begin{rawpromptbox}
five bears and a donut.
\end{rawpromptbox}
\begin{enhancedpromptbox}
A group of five small, plush bears is arranged in a circle on a plain, light-colored surface, with a single large donut placed among them. The donut, positioned in the center, is perfectly round with a golden-brown, fried surface and is covered in a thick, rainbow-colored glaze topped with colorful sprinkles. Surrounding the donut are the five bears, each with a distinct color and texture; one brown bear holds the donut, while a black bear, a beige bear, and two other brown bears look towards it. The bears are made of a fluffy, plush material, featuring simple stitched noses and button eyes. This image presents a photography style with a shallow depth of field, emphasizing the subjects.
\end{enhancedpromptbox}
\caption{\textbf{Comparison between the raw prompt and the enhanced prompt.} Here, we use \texttt{PromptEnhancer-32B} as PE.}
\label{fig:prompt-comparison3}
\end{figure*}

\section{Additional Benchmark Results}\label{app:benchresult}
\subsection{GenEval Details}
In Table~\ref{app:tab_geneval} and Table~\ref{app:tab_geneval_other}, we provide the detailed GenEval performance breakdown for different models and PEs. In particular, we report the fine-grained performance in Single Object, Two Object, Counting, Colors, Position, and Attribute Binding. We also report the Overall score.
\begin{table}[h]
    \centering
    \caption{\textbf{Detailed GenEval performance breakdown.} We report the fine-grained performance in Single Object, Two Object, Counting, Colors, Position, and Attribute Binding. We also report the Overall score. Here, we use \texttt{GPT-5.6 Sol} as PE.}\label{app:tab_geneval}
    \setlength{\tabcolsep}{1.5mm}
    \small
    \begin{tabular}{lcccccc|c}
    \toprule
    \textbf{Method} & \textbf{Single Obj} & \textbf{Two Obj} & \textbf{Counting} & \textbf{Color} &\textbf{Position} & \textbf{Attr Binding} & \textbf{Overall} \\
    \midrule
    \rowcolor{lightgraycell}
    \multicolumn{8}{c}{\textbf{\texttt{SD3.5-M (2.5B)}}} \\
    Base & 0.975 &0.778 &0.613 &0.787 &0.223 &0.472 &0.628\\
    Base+PE & 0.959 & 0.838& 0.634& 0.838& 0.603& 0.635&0.743\\
    SFT & 0.972 &0.856 &0.691 &0.832 &0.512 &0.500 &0.718\\
    Off-Policy Distillation &0.981 &0.846 &0.628 &0.838 &0.698 &0.650 &0.770\\
    \rowcolor{lightbluecell}
    \textbf{PE-OPSD (Ours)} &0.988 &0.886 &0.631 &0.878 &0.698 &0.710 &0.797\\
    \midrule
    \rowcolor{lightgraycell}
    \multicolumn{8}{c}{\textbf{\texttt{Z-Image (6B)}}} \\
    Base &0.959 &0.801 &0.569 &0.809 &0.338 &0.480 &0.650\\
    Base+PE &0.978& 0.879& 0.619& 0.888& 0.755& 0.743&0.810\\
    SFT &0.972 &0.884 &0.697 &0.902 &0.780 &0.698 &0.821\\
    Off-Policy Distillation &0.988 &0.886 &0.641 &0.899 &0.762 &0.767 &0.824\\
    \rowcolor{lightbluecell}
    \textbf{PE-OPSD (Ours)} & 0.988 &0.884 &0.656 &0.918 &0.823 &0.805 &0.846\\
    \midrule
    \rowcolor{lightgraycell}
    \multicolumn{8}{c}{\textbf{\texttt{Z-Image-Turbo (6B)}}} \\
    Base &0.988 &0.833 &0.759 &0.859 &0.460 &0.588 &0.737\\
    Base+PE &0.975&0.864&0.803&0.923&0.770&0.787&0.850\\
    SFT &0.981 &0.732 &0.625 &0.894 &0.708 &0.677 &0.766\\
    Off-Policy Distillation &0.984 &0.866 &0.778 &0.907 &0.805 &0.777 &0.851\\
    \rowcolor{lightbluecell}
    \textbf{PE-OPSD (Ours)} &0.969 &0.879 &0.794 &0.915 &0.797 &0.835 &0.863\\
    \midrule
    \rowcolor{lightgraycell}
    \multicolumn{8}{c}{\textbf{\texttt{FLUX.2-klein-base (9B)}}} \\
    Base &0.994 &0.851 &0.706 &0.904 &0.632 &0.603 &0.775\\
    Base+PE &0.984& 0.904& 0.797& 0.928& 0.820& 0.757&0.862\\
    \rowcolor{lightbluecell}
    \textbf{PE-OPSD (Ours)} &0.991 &0.904 &0.819 &0.912 &0.840 &0.785 &0.873\\
    \midrule
    \rowcolor{lightgraycell}
    \multicolumn{8}{c}{\textbf{\texttt{FLUX.2-klein (9B)}}} \\
    Base &0.994 &0.912 &0.847 &0.894 &0.733 &0.787 &0.856\\
    Base+PE &0.991& 0.896& 0.775& 0.912& 0.835& 0.757& 0.859\\
    \rowcolor{lightbluecell}
    \textbf{PE-OPSD (Ours)} &0.988 &0.874 &0.747 &0.931 &0.845 &0.823 &0.867\\
    \midrule
    \rowcolor{lightgraycell}
    \multicolumn{8}{c}{\textbf{\texttt{QwenImage-2512 (20B)}}} \\
    Base &0.984 &0.798 &0.331 &0.832 &0.302 &0.500 &0.620\\
    Base+PE &0.975&0.904&0.694&0.891&0.755&0.818&0.839\\
    \rowcolor{lightbluecell}
    \textbf{PE-OPSD (Ours)} &0.981 &0.899 &0.706 &0.910 &0.810 &0.833 &0.857\\
    \bottomrule
    \end{tabular}
\end{table}

\begin{table}[h]
    \centering
    \caption{\textbf{Detailed GenEval performance breakdown on Z-image with other PEs.} We report the fine-grained performance in Single Object, Two Object, Counting, Colors, Position, and Attribute Binding. We also report the Overall score.}\label{app:tab_geneval_other}
    \setlength{\tabcolsep}{1.5mm}
    \small
    \begin{tabular}{lcccccc|c}
    \toprule
    \textbf{Method} & \textbf{Single Obj} & \textbf{Two Obj} & \textbf{Counting} & \textbf{Color} &\textbf{Position} & \textbf{Attr Binding} & \textbf{Overall} \\
    \midrule
    \rowcolor{lightgraycell}
    \multicolumn{8}{c}{\textbf{\texttt{PromptEnhancer-7B as PE}}} \\
    Base &0.959 &0.801 &0.569 &0.809 &0.338 &0.480 &0.650\\
    Base+PE &0.947&0.801&0.588&0.750&0.505&0.555&0.684\\
    SFT &0.966 &0.871 &0.762 &0.835 &0.627 &0.560 &0.763\\
    Off-Policy Distillation &0.978 &0.856 &0.716 &0.856 &0.613 &0.623 &0.767\\
    \rowcolor{lightbluecell}
    \textbf{PE-OPSD (Ours)} &0.978 &0.864 &0.647 &0.872 &0.610 &0.740 &0.782\\
    \midrule
    \rowcolor{lightgraycell}
    \multicolumn{8}{c}{\textbf{\texttt{PromptEnhancer-32B as PE}}} \\
    Base &0.959 &0.801 &0.569 &0.809 &0.338 &0.480 &0.650\\
    Base+PE &0.984&0.854&0.675&0.902&0.797&0.690&0.815\\
    SFT &0.972 &0.884 &0.741 &0.888 &0.825 &0.680 &0.829\\
    Off-Policy Distillation &0.981 &0.889 &0.725 &0.896 &0.853 &0.718 &0.842\\
    \rowcolor{lightbluecell}
    \textbf{PE-OPSD (Ours)} &0.991 &0.904 &0.666 &0.941 &0.882 &0.810 &0.868\\
    \bottomrule
    \end{tabular}
\end{table}

\subsection{GenEval2 Details}
In Table~\ref{app:tab_geneval2} and Table~\ref{app:tab_geneval2_other}, we provide the detailed GenEval2 performance breakdown for different models and PEs. In particular, we report the fine-grained performance in Object, Attribute, Count, Position, and Verb. We also report the overall soft-tifa score in GenEval2$_{\texttt{AM}}$ and GenEval2$_{\texttt{GM}}$.
\begin{table}[h]
    \centering
    \caption{\textbf{Detailed GenEval2 performance breakdown.} We report the fine-grained performance in Object, Attribute, Count, Position, and Verb. We also report the overall soft-tifa score in GenEval2$_{\texttt{AM}}$ and GenEval2$_{\texttt{GM}}$. Here, we use \texttt{GPT-5.6 Sol} as PE.}\label{app:tab_geneval2}
    \setlength{\tabcolsep}{1.8mm}
    \small
    \begin{tabular}{lccccc|cc}
    \toprule
    \textbf{Method} & \textbf{Object} & \textbf{Attribute} & \textbf{Count} & \textbf{Position} & \textbf{Verb} & \textbf{GenEval2$_{\texttt{AM}}$} & \textbf{GenEval2$_{\texttt{GM}}$} \\
    \midrule
    \rowcolor{lightgraycell}
    \multicolumn{8}{c}{\textbf{\texttt{SD3.5-M (2.5B)}}} \\
    Base &0.859 &0.672 &0.431 &0.397 &0.161 &0.633 &0.176\\
    Base+PE &0.868& 0.751& 0.476& 0.521& 0.207 &0.680 &0.220\\
    SFT &0.835 &0.703 &0.478 &0.438 &0.124 &0.656 &0.207\\
    Off-Policy Distillation &0.850 &0.715 &0.461 &0.483 &0.195& 0.671 &0.223\\
    \rowcolor{lightbluecell}
    \textbf{PE-OPSD (Ours)} &0.868 &0.758 &0.477 &0.489 &0.179 &0.682 &0.226\\
    \midrule
    \rowcolor{lightgraycell}
    \multicolumn{8}{c}{\textbf{\texttt{Z-Image (6B)}}} \\
    Base &0.927 &0.844 &0.599 &0.607 &0.294 &0.761 &0.306\\
    Base+PE &0.973&0.923&0.637&0.826&0.459&0.827 &0.404\\
    SFT &0.981 &0.906 &0.643 &0.816 &0.323 &0.834 &0.403\\
    Off-Policy Distillation &0.975 &0.931 &0.625 &0.814 &0.320 &0.828 &0.395\\
    \rowcolor{lightbluecell}
    \textbf{PE-OPSD (Ours)} & 0.975 &0.938 &0.634 &0.816 &0.381 &0.831 &0.406\\
    \midrule
    \rowcolor{lightgraycell}
    \multicolumn{8}{c}{\textbf{\texttt{Z-Image-Turbo (6B)}}} \\
    Base &0.970 &0.763 &0.665 &0.624 &0.196 &0.783 &0.341\\
    Base+PE &0.973&0.887&0.688&0.845&0.285 &0.843 &0.479\\
    SFT &0.929 &0.866 &0.601 &0.799 &0.205 &0.788 &0.413\\
    Off-Policy Distillation &0.968 &0.892 &0.640 &0.821 &0.297 &0.821 &0.437\\
    \rowcolor{lightbluecell}
    \textbf{PE-OPSD (Ours)} &0.975 &0.894 &0.686 &0.851 &0.353 &0.844 &0.501\\
    \midrule
    \rowcolor{lightgraycell}
    \multicolumn{8}{c}{\textbf{\texttt{FLUX.2-klein-base (9B)}}} \\
    Base &0.929 &0.863 &0.567 &0.710 &0.340 &0.773 &0.359\\
    Base+PE &0.962& 0.950&0.647&0.838&0.477&0.838 &0.442\\
    \rowcolor{lightbluecell}
    \textbf{PE-OPSD (Ours)} &0.956 &0.945 &0.641 &0.830 &0.399 &0.831 &0.444\\
    \midrule
    \rowcolor{lightgraycell}
    \multicolumn{8}{c}{\textbf{\texttt{FLUX.2-klein (9B)}}} \\
    Base &0.954 &0.889 &0.593 &0.742 &0.315 &0.797 &0.348\\
    Base+PE &0.961&0.952&0.609&0.843&0.421&0.827&0.381\\
    \rowcolor{lightbluecell}
    \textbf{PE-OPSD (Ours)} &0.960 &0.947 &0.616 &0.846 &0.362 &0.825 &0.434\\
    \midrule
    \rowcolor{lightgraycell}
    \multicolumn{8}{c}{\textbf{\texttt{QwenImage-2512 (20B)}}} \\
    Base &0.930 &0.593 &0.526 &0.510 &0.270 &0.671 &0.163\\
    Base+PE &0.977&0.848&0.620 &0.843&0.423 &0.813&0.380\\
    \rowcolor{lightbluecell}
    \textbf{PE-OPSD (Ours)} &0.978 &0.882 &0.625 &0.851 &0.397 &0.820 &0.399\\
    \bottomrule
    \end{tabular}
\end{table}

\begin{table}[t]
    \centering
    \caption{\textbf{Detailed GenEval2 performance breakdown on Z-Image with other PEs.} We report the fine-grained performance in Object, Attribute, Count, Position, and Verb. We also report the overall soft-tifa score in GenEval2$_{\texttt{AM}}$ and GenEval2$_{\texttt{GM}}$.}\label{app:tab_geneval2_other}
    \setlength{\tabcolsep}{1.8mm}
    \small
    \begin{tabular}{lccccc|cc}
    \toprule
    \textbf{Method} & \textbf{Object} & \textbf{Attribute} & \textbf{Count} & \textbf{Position} & \textbf{Verb} & \textbf{GenEval2$_{\texttt{AM}}$} & \textbf{GenEval2$_{\texttt{GM}}$} \\
    \midrule
    \rowcolor{lightgraycell}
    \multicolumn{8}{c}{\textbf{\texttt{PromptEnhancer-7B as PE}}} \\
    Base &0.927 &0.844 &0.599 &0.607 &0.294 &0.761 &0.306\\
    Base+PE &0.962&0.818&0.653&0.723&0.471&0.801&0.375\\
    SFT &0.975 &0.865 &0.660 &0.765 &0.340 &0.819 &0.400\\
    Off-Policy Distillation &0.975 &0.832 &0.649 &0.720 &0.274 &0.804 &0.380\\
    \rowcolor{lightbluecell}
    \textbf{PE-OPSD (Ours)} &0.979 &0.875 &0.636 &0.784 &0.388 &0.819 &0.422\\
    \midrule
    \rowcolor{lightgraycell}
    \multicolumn{8}{c}{\textbf{\texttt{PromptEnhancer-32B as PE}}} \\
    Base &0.927 &0.844 &0.599 &0.607 &0.294 &0.761 &0.306\\
    Base+PE &0.966&0.924&0.664&0.864&0.543&0.844 &0.470\\
    SFT &0.981 &0.925 &0.672 &0.848 &0.339 &0.845 &0.440\\
    Off-Policy Distillation &0.970 &0.940 &0.662 &0.832 &0.382 &0.843 &0.459\\
    \rowcolor{lightbluecell}
    \textbf{PE-OPSD (Ours)} &0.982 &0.952 &0.655 &0.845 &0.430 &0.846 &0.476\\
    \bottomrule
    \end{tabular}
\end{table}

\clearpage
\subsection{EvalMuse Details}
In Table~\ref{app:tab_evalmuse}, we provide the detailed performance breakdown for different models. In particular, we report the fine-grained performance in Attribute, Location, Color, Object, Material, A./H. (Animal/Human), Food, Shape, Activity, Spatial, and Counting. We also report the overall score following the official guidance~\citep{han2026evalmuse}.

\begin{table}[h]
    \centering
    \caption{\textbf{Detailed EvalMuse performance breakdown.} We report the fine-grained performance in Attribute, Location, Color, Object, Material, A./H. (Animal/Human), Food, Shape, Activity, Spatial, and Counting. We also report the overall score. \textbf{Bold:} best in Overall.}\label{app:tab_evalmuse}
    \setlength{\tabcolsep}{1mm}
    \scriptsize
    \begin{tabular}{lccccccccccc|c}
    \toprule
    \textbf{Method} & \textbf{Attribute} & \textbf{Location} & \textbf{Color} & \textbf{Object} & \textbf{Material} & \textbf{A./H.} & \textbf{Food} & \textbf{Shape} & \textbf{Activity} & \textbf{Spatial} & \textbf{Counting} & \textbf{Overall} \\
    \midrule
    SD3.5-M&0.801 &0.724 &0.570 &0.675 &0.542 &0.524 &0.654 &0.826 &0.539 &0.459 &0.234&3.199\\
    \rowcolor{lightbluecell}
    \qquad +Ours &0.820 &0.763 &0.634 &0.710 &0.587 &0.562 &0.697 &0.832 &0.551 &0.588 &0.283 & \textbf{3.368}\\
    \midrule
    Z-Image &0.797 &0.723 &0.618 &0.709 &0.620 &0.608 &0.656 &0.806 &0.582 &0.655 &0.314 &3.340\\
    \rowcolor{lightbluecell}
    \qquad +Ours &0.809 &0.768 &0.694 &0.738 &0.681 &0.654 &0.716 &0.811 &0.596 &0.643 &0.384 &\textbf{3.561}\\
    \midrule
    Z-Image-Turbo &0.800 &0.762 &0.711 &0.727 &0.707 &0.620 &0.699 &0.798 &0.589 &0.641 &0.340 &3.515\\
    \rowcolor{lightbluecell}
    \qquad +Ours &0.807 &0.753 &0.626 &0.729 &0.639 &0.632 &0.735 &0.806 &0.620 &0.584 &0.337 &\textbf{3.534}\\
    \midrule
    FLUX.2-klein-base &0.834 &0.784 &0.707 &0.749 &0.680 &0.659 &0.708 &0.765 &0.644 &0.661 &0.311 &3.638\\
    \rowcolor{lightbluecell}
    \qquad +Ours &0.829 &0.797 &0.728 &0.755 &0.691 &0.700 &0.733 &0.798 &0.664 &0.680 &0.367 &\textbf{3.733}\\
    \midrule
    FLUX.2-klein &0.832 &0.778 &0.716 &0.761 &0.686 &0.691 &0.728 &0.834 &0.661 &0.675 &0.337 &\textbf{3.724}\\
    \rowcolor{lightbluecell}
    \qquad +Ours &0.832 &0.788 &0.725 &0.760 &0.691 &0.687 &0.726 &0.810 &0.660 &0.686 &0.333 &3.710\\
    \midrule
    QwenImage-2512 &0.817 &0.757 &0.616 &0.720 &0.662 &0.685 &0.708 &0.835 &0.618 &0.605 &0.280 &3.485\\
    \rowcolor{lightbluecell}
    \qquad +Ours &0.829 &0.798 &0.749 &0.772 &0.750 &0.693 &0.736 &0.837 &0.643 &0.725 &0.397 &\textbf{3.756}\\
    \bottomrule
    \end{tabular}
\end{table}

% \subsection{Additional Out-of-domain benchmark results}
% \begin{table}[!t]
% \centering
% \caption{\textbf{Results on out-of-domain benchmarks.} For DPG-bench, we report Global (Glo.), Entity (Ent.), Attribute (Attr.), Relation (Rela.), Other and Overall$^\star$. For T2I-CompBench++, we report Color, Shape, Textual (Tex.), Numeracy (Num.), Complex (Comp.), Spatial (Spa.), 3D Spatial (3D Spa.), Non-spatial (Non-spa.) and Overall$^\star$. \textbf{Bold:} best in Overall$^\star$.}
% {
% \tiny
% \setlength{\tabcolsep}{1mm}
% \renewcommand{\arraystretch}{0.9}
% \begin{tabular}{lcccccccccccccccc}
% \toprule
% \multirow{2}{*}[-0.8ex]{\textbf{Method}}
% & \multicolumn{6}{c}{\textbf{DPG-bench}}
% & \multicolumn{9}{c}{\textbf{T2I-CompBench++}}
% \\
% \cmidrule(lr){2-7}
% \cmidrule(lr){8-16}
% & \textbf{Overall$^\star$}
% & \textbf{Glo.}
% & \textbf{Ent.}
% & \textbf{Attr.}
% & \textbf{Rela.}
% & \textbf{Other}
% & \textbf{Overall$^\star$}
% & \textbf{Color}
% & \textbf{Shape}
% & \textbf{Tex.}
% & \textbf{Num.}
% & \textbf{Comp.}
% & \textbf{Spa.}
% & \textbf{3D Spa.}
% & \textbf{Non-spa.}
% \\
% \midrule
% FLUX.2-klein-base
% & -\\
% \rowcolor{lightbluecell}
% \qquad +Ours
% & -\\
% \bottomrule
% \end{tabular}
% }
% \label{tab:app_benchmark}
% \end{table}

\section{System Prompts}

\subsection{System Prompt for Consistency Filtering}
\label{app:sp1}

For GPT-generated enhanced prompts, we employ an consistency judge to identify expansions that alter or omit critical semantics from the original prompt. The judge verifies exact counts, object identities, attributes, actions, and spatial relations, and returns a binary decision in a structured JSON format. The exact system prompt is provided below.

\begin{tcolorbox}[
    colback=gray!3,
    colframe=gray!45,
    boxrule=0.5pt,
    arc=1pt,
    left=6pt,
    right=6pt,
    top=6pt,
    bottom=6pt,
    title=\textbf{System prompt for consistency filtering},
    fonttitle=\small,
    fontupper=\scriptsize
]
You are a strict consistency judge for enhanced text-to-image prompts.

\medskip
The original prompt is the source of truth. The candidate enhanced prompt is consistent
only if it explicitly preserves every critical fact:

\begin{itemize}[leftmargin=20pt, nosep]
    \item exact counts and object identities;
    \item colors, visual material appearances, textures, and other attributes;
    \item actions and subject/object roles;
    \item spatial relations and ordering.
\end{itemize}

\medskip
Do not treat a fact as preserved when it is merely implied. Synonyms and grammatical number words are acceptable when their meaning is unambiguous. Additional composition, lighting, style, and background details are acceptable only when they do not alter or contradict an original fact. Be especially strict about swapped roles, changed counts, merged object groups, conflicting attributes, and missing relations.

\medskip
Do not relax colors, counts, identities, actions, or relations. In particular, spots, reflections, or small localized details do not satisfy an overall color requirement when the object is explicitly described as a different base color.

\medskip
Return exactly one JSON object with no Markdown:

\texttt{\{"consistent": true, "issues": []\}}

or

\texttt{\{"consistent": false, "issues": ["specific issue", "..."]\}}

\medskip
Judge only. Do not rewrite, complete, or suggest a replacement prompt.
\end{tcolorbox}

\clearpage
\subsection{System prompts for generating enhanced prompts.}\label{app:sp}
we provide the system prompt used by \texttt{GPT-5.6 Sol} to generate enhanced prompts. Given a raw prompt, the enhancer is instructed to produce only the expanded caption without auxiliary explanations or Markdown formatting. The instruction encourages a hierarchical description while explicitly preserving the objects, attributes, spatial relations, rendered text, and intellectual-property subjects specified in the original prompt.

\begin{tcolorbox}[
    colback=gray!3,
    colframe=gray!45,
    boxrule=0.5pt,
    arc=1pt,
    left=6pt,
    right=6pt,
    top=6pt,
    bottom=6pt,
    title=\textbf{System prompt for GPT-5.6 Sol},
    fonttitle=\small,
    fontupper=\scriptsize
]
You are an expert in writing prompts for image generation. I will give you a sentence, and you are to expand this sentence into a detailed caption for generating an image. And the captions must follow the rules listed below.

\medskip
\textbf{I. Sentence Structures}
The captions follow a consistent, hierarchical structure that moves from a general overview to specific details.

\begin{enumerate}[leftmargin=20pt, nosep]
    \item The Opening Statement: General Overview
    \item The Body: Systematic and Spatially Organized Description
    \item Hierarchical Object Description: From Whole to Parts
    \item The Concluding Statement: Stylistic Identification
\end{enumerate}

\medskip
\textbf{II. Grammatical Rules}
The grammar is precise, descriptive, and maintains an objective tone.

\begin{enumerate}[leftmargin=20pt, nosep]
    \item Tense: Consistent Present Tense
    \item Voice: Mix of Active and Passive
    \item Prepositional Phrases for Precision
    \item Participial Phrases for Efficient Detail
    \item Rich and Specific Adjectives
    \item Precision and Hedging Language
    \item Complex and Compound Sentences
\end{enumerate}

\medskip
\textbf{Key constraints:}
\begin{enumerate}[leftmargin=20pt, nosep]
    \item Only provide the final captions, do not use markdown format.
    \item The expanded captions must follow the rules listed above.
    \item The expanded captions should adhere to the original sentence, especially the subject and the subject's attributes, including color, size, spatial relationships, etc.
    \item You can use your world knowledge to expand some professional terminology to proper explanations that suitable for image generation models.
    \item If the style of original sentence is not mentioned, you should assume it is a photography style. And you can infer the style from the context of the sentence if the photography style is not suitable.
    \item Describe the scene or subject directly, do not use ``The image'', ``The composition'', ``The scene'' and similar words in the beginning of the captions.
    \item If the original sentence has a IP subject, you should keep the IP subject in the expanded captions, and describe the background of the IP in the expanded captions.
    \item If the original sentence has a text that need to be rendered, you should keep the text in the expanded captions, and format text as ``rendered text''.
\end{enumerate}

\medskip
Next, I will give you my sentence. Please provide the expanded captions:
\end{tcolorbox}

\section{Additional Qualitative Examples}\label{app:example}
\begin{figure}[h]
    \centering
    \includegraphics[width=1.0\linewidth]{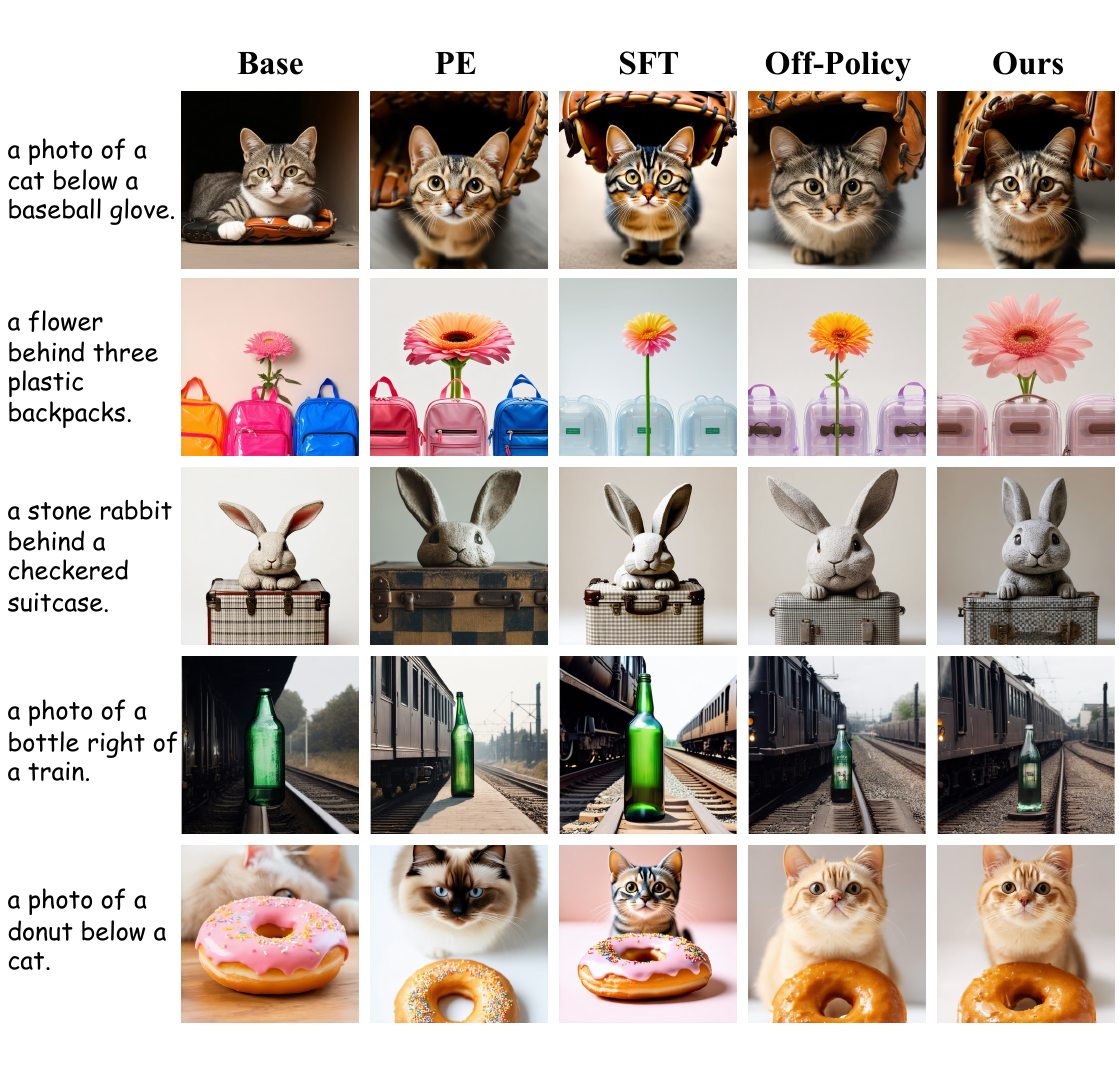}
    \caption{\textbf{Visual examples in SD3.5-M.} We use the \texttt{GPT-5.6 Sol} as PE.}
    \label{fig:vis_sd}
\end{figure}

\begin{figure}[h]
    \centering
    \includegraphics[width=1.0\linewidth]{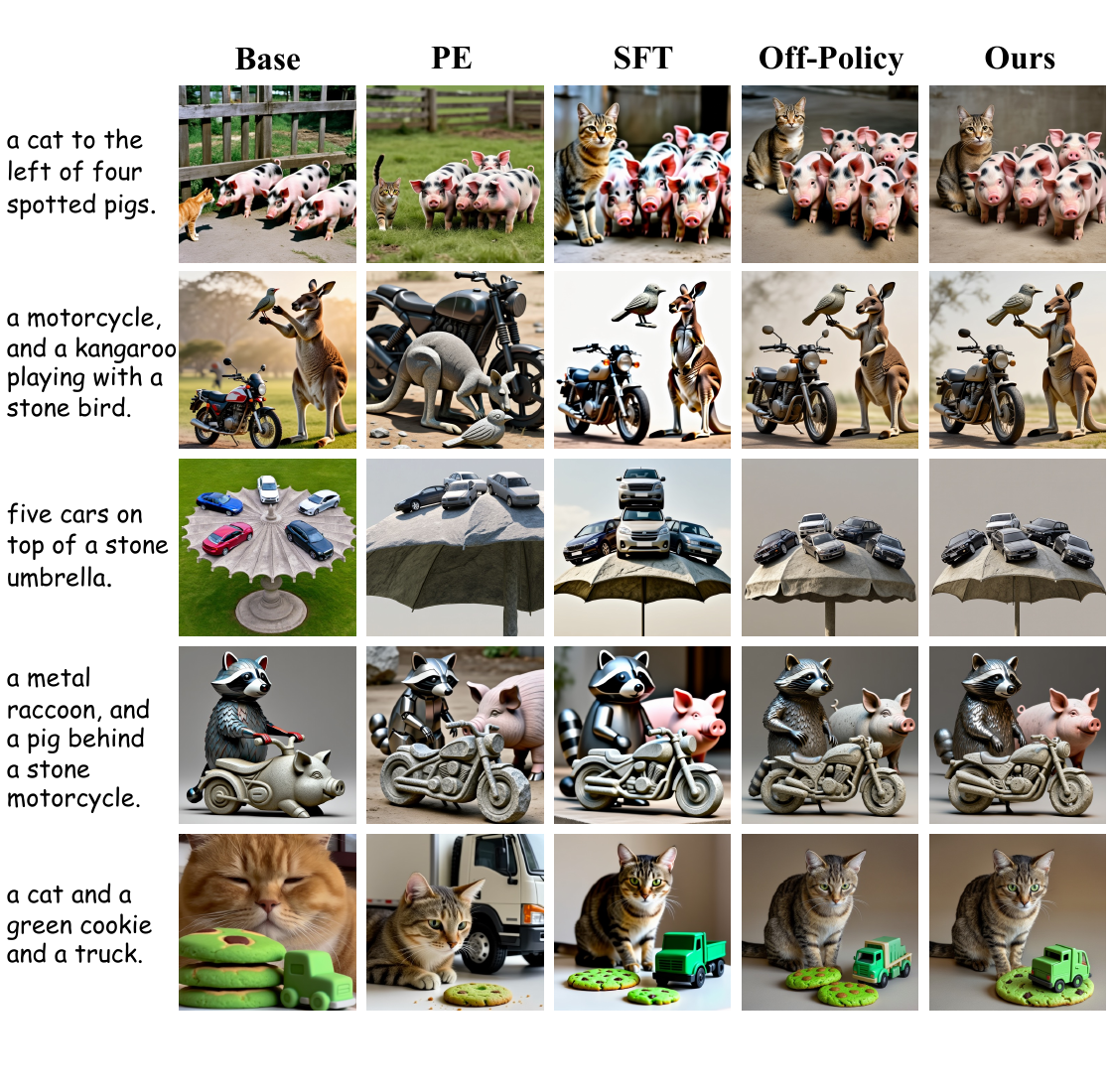}
    \caption{\textbf{Visual examples in Z-Image.} We use the \texttt{GPT-5.6 Sol} as PE.}
    \label{fig:vis_z}
\end{figure}

\begin{figure}[h]
    \centering
    \includegraphics[width=1.0\linewidth]{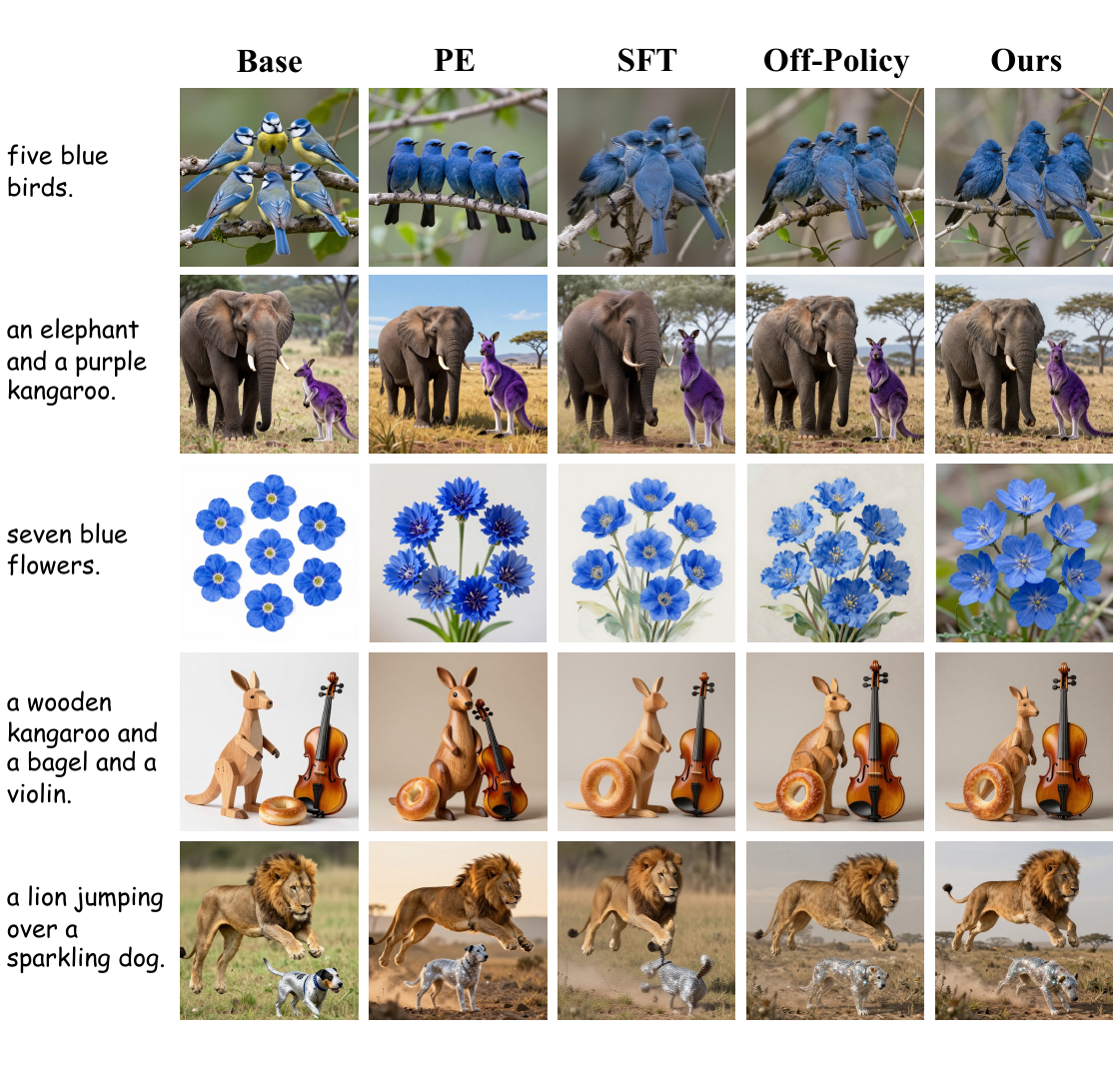}
    \caption{\textbf{Visual examples in Z-Image-Turbo.} We use the \texttt{GPT-5.6 Sol} as PE.}
    \label{fig:vis_zturbo}
\end{figure}

We provide more visual examples across different models and methods in Figure~\ref{fig:vis_sd}--Figure~\ref{fig:vis_zturbo}.

\section{Limitations and Future Work}
PE-OPSD uses enhanced prompts as privileged supervision, and its learning signal is therefore influenced by the quality and semantic faithfulness of the selected PE. Although our consistency filtering and experiments with multiple PEs demonstrate robust improvements, inaccurate or overly specific rewrites may still introduce undesirable supervision. Future work could investigate confidence-aware filtering, agreement across multiple PEs, and adaptive supervision that emphasizes reliable prompt elements.

PE-OPSD shifts computation from per-request inference to offline prompt construction and post-training. Reducing this cost through selective timestep supervision and fewer teacher evaluations is a promising direction. More broadly, extending privileged-condition distillation beyond flow-matching text-to-image models may further establish its applicability across generative paradigms.

\end{document}